\documentclass{bmvc2k}

\usepackage{hhline}
\usepackage{color, colortbl}
\usepackage{multirow}
\usepackage{times}
\usepackage{epsfig}
\usepackage{graphicx}
\usepackage{amsmath}
\usepackage{amssymb}
\usepackage[utf8]{inputenc}
\usepackage{lipsum} 
\usepackage{booktabs}
\usepackage{makecell}
\newcommand{\ra}[1]{\renewcommand{\arraystretch}{#1}}

\title{Is Face Recognition Sexist? 
\\No, Gendered Hairstyles and Biology Are
}

\addauthor{Vítor Albiero}{valbiero@nd.edu}{1}
\addauthor{Kevin W. Bowyer}{kwb@nd.edu}{1}

\addinstitution{
 Computer Vision Research Lab\\
 University of Notre Dame\\
 Notre Dame, USA
}

\runninghead{Albiero and Bowyer}{Is Face Recognition Sexist?}
\begin{document}

\maketitle

\begin{abstract}
Recent news articles have accused face recognition of being ``biased'', ``sexist'' or ``racist''.
There is consensus in the research literature that face recognition accuracy is lower for females, who often have both a higher 
false match rate and a higher false non-match rate.
However, there is little published research aimed at identifying the cause of lower accuracy for females.
For instance, the 2019 Face Recognition Vendor Test that documents lower female accuracy across a broad range of algorithms and datasets also lists
``Analyze cause and effect’’ under the heading ``What we did not do’’.
We present the first experimental analysis to identify major causes of lower face recognition accuracy for females on datasets where previous research has observed this result.
Controlling for equal amount of visible face in the test images reverses the apparent higher false non-match rate for females. 
Also, principal component analysis indicates that images of two different females are inherently more similar than of two different males, potentially accounting for a difference in false match rates.

\end{abstract}

\section{Introduction}

Many news articles in recent years have criticized face recognition 
as being ``biased'', ``sexist'' or ``racist'' \cite{Lohr2018,Hoggins2019,Doctorow2019,Santow2020}.
Various papers summarized in Related Work have reported
that face recognition is less accurate for females.
Accuracy also varies between racial groups, and age ranges, but this paper focuses on the female / male difference.
Surprisingly, 
given the media attention on the issue, 
the underlying cause(s) of lower accuracy for females are heretofore unknown.
The popularity of deep CNN algorithms perhaps naturally gives rise to the common 
speculation that the cause is female under-representation in the training data.
But recent studies show that simply using a gender-balanced training set does not result in balanced accuracy on test data \cite{Albiero2020_train, Albiero2020_gender}. 

This paper is the first to present experimental results that
explain the causes of lower face recognition accuracy for females,
on datasets where this result is observed in previous research.
To promote reproducible and transparent results, we use a state-of-the-art deep CNN matcher and datasets that are available to other researchers.
Our experiments show that gendered hairstyles result in, on average, more of the face being occluded for females, and that the inherent variability between different female faces is lower.
%
%

\section{Related Work}
\label{sec:related_work}

Drozdowski et al. \cite{Drozdowski2020} give a current, broad survey 
on demographic bias in biometrics. Here we focus on selected prior works dealing specifically with female / male accuracy difference.

The earliest work we are aware of to report lower accuracy for females is the 2002 Face Recognition Vendor Test (FRVT) \cite{frvt}. 
Evaluating ten algorithms of that era, 
identification rates of the top systems are 6\% to 9\% higher for males.
However, for the highest-accuracy matcher, accuracy was essentially the same. 
In a 2009 meta-analysis of results from eight prior papers, Lui et al. \cite{Lui2009} 
concluded that there was a weak pattern of lower accuracy for females, and noted interactions with factors such as age, expression, lighting and indoor/outdoor imaging.
Beveridge et al. \cite{Beveridge2009} analyzed results for three algorithms on a  Face Recognition Grand Challenge \cite{frgc} dataset, and found that males had a higher verification rate for the two higher-accuracy algorithms and females had a higher verification rate for the third, lower-accuracy algorithm. 
Klare et al \cite{Klare2012} presented results from three commercial off-the-shelf (COTS) and three research algorithms showing that females had a worse receiver operating characteristic (ROC) curve for all six, and also showed example results for which females had both a worse impostor distribution and a worse genuine distribution.

The above works are from before the 
deep convolutional neural network (CNN) wave in face recognition. 
Related works since the rise of deep CNN matchers report similar results. 

Cook et al. \cite{cook2018} analyze images acquired using eleven different automated kiosks with the same COTS matcher, and report that genuine scores are lower for females.
Lu et al. \cite{Lu2018} use deep CNN matchers and datasets from the IARPA Janus program in a detailed study of the effect of various covariates, and report that accuracy is lower for females.
Howard et al. \cite{bio_rally}
report higher false match rates for African-American and Caucasian females under age 40. For subjects over 40, African-American females again had a higher false match rate, but Caucasian females had a lower false match rate than Caucasian males. However, the dataset is the smallest of the recent papers on this topic, with just 363 subjects, divided into 8 subsets for this analysis.
Vera-Rodriguez et al. \cite{Vera-Rodriguez2019} use a Resnet-50 and a VGGFace matcher with the VGGFace2 dataset \cite{vggface2}, and report that females have a worse ROC for both.
Albiero et al. \cite{Albiero2020_gender} use the ArcFace matcher with four datasets,
and report that the general pattern of results is that females have a worse ROC curve, impostor distribution and genuine distribution.
Similarly, Krishnapriya et al. \cite{KrishnapriyaTTS} show that both the ArcFace and the VGGFace2 matcher, which are trained on different datasets and with different loss functions, result in a worse ROC curve, impostor distribution and genuine distribution for females, for both African-American and Caucasian image cohorts.
Grother et al. \cite{frvt3}, in the 2019 FRVT focused on demographic analysis, found that females have higher false match rates (worse impostor distribution), and that the phenomenon is consistent across a wide range of different matchers and datasets. 
Relative to the genuine distribution, they report that women often have higher false negative error rates, but also note that there are exceptions to this generalization.

The general pattern across these works, for a broad variety of matchers and datasets, is that face recognition accuracy is lower for females, with females having a worse impostor distribution and a worse genuine distribution.

Several researchers have speculated cosmetics use as a cause of the gender gap in accuracy \cite{Klare2012, cook2018, Lu2018}.
Dantcheva et al. \cite{can_facial_cosmetics} and others have documented how the use of cosmetics can degrade face recognition accuracy.
In scenarios where females have substantial cosmetics use and males do not, this is a possible cause of the accuracy gender gap.
However, Albiero et al \cite{Albiero2020_gender} showed that cosmetics plays at most a minor role in the MORPH dataset, which is the main dataset also used in this paper.
Other speculated causes include more varied hairstyles for females \cite{Albiero2020_gender} and shorter height for women, leading to non-optimal camera angle \cite{Grother2010, cook2018}.

Buolamwini and Gebru \cite{gendershades}, Muthukumar et al. \cite{Muthukumar} and others have looked at the accuracy of algorithms that predict gender from a face image, and found lower accuracy for females.
Gender classification of a face image is a related problem but distinct from face recognition.

Very few works attempt to identify the cause(s) of the accuracy gender gap.
For example, the recent NIST report on demographic effects \cite{frvt3} lists ``analyze cause and effect’’ as an item under ``what we did not do’’.
However, Albiero et al. \cite{Albiero2020_gender} reported on experiments to determine if cosmetics use, facial expression, forehead occlusion by hair, or downward-looking camera angle could explain the worse impostor and genuine distributions observed for females.
They found that these factors, either individually or together, did not significantly impact the gender gap in accuracy.
A popular speculation is that accuracy is lower for females  because they are under-represented in the training data. Albiero et al. \cite{Albiero2020_train} trained matchers from scratch using male / female ratios of 0/100, 25/75, 50/50, 75/25 and 100/0.
They found that explicitly gender-balanced training data does not result in balanced accuracy on test data, 
%
%
and that in fact a 25\% male / 75\% female training set resulted in the least-imbalanced test accuracy, 
but was generally not the training mix that maximized female, male, or average accuracy.


\begin{figure*}[t]
  \begin{subfigure}[b]{1\linewidth}
      \begin{subfigure}[b]{0.327\linewidth}
        \centering
          \includegraphics[width=\linewidth]{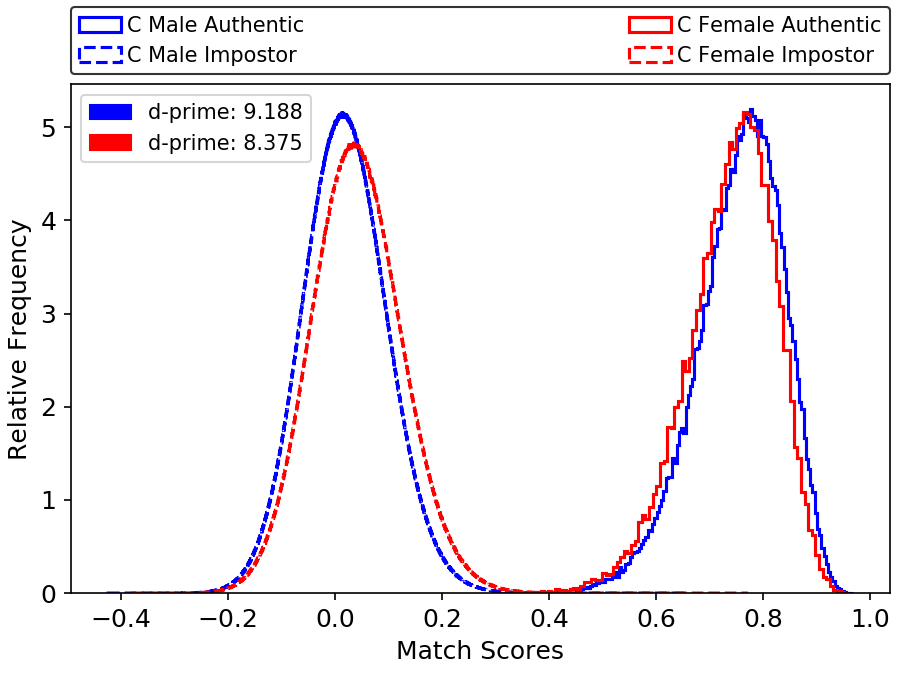}
          \caption{MORPH Caucasian}
      \end{subfigure}
      \hfill 
      \begin{subfigure}[b]{0.327\linewidth}
        \centering
          \includegraphics[width=\linewidth]{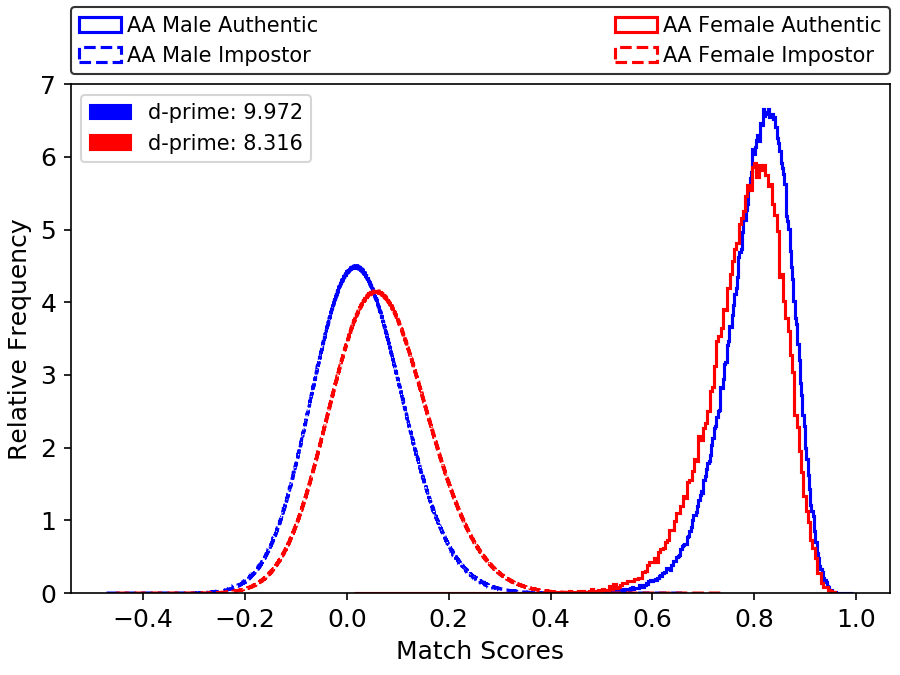}
          \caption{MORPH African-American}
      \end{subfigure}
      \hfill 
      \begin{subfigure}[b]{0.327\linewidth}
        \centering
          \includegraphics[width=\linewidth]{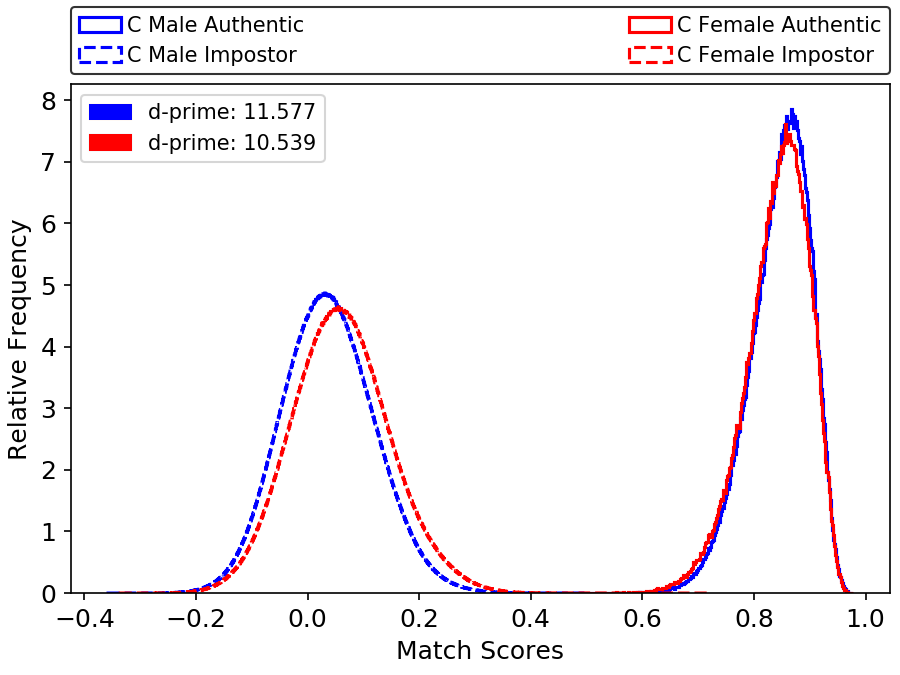}
          \caption{Notre Dame Caucasian}
      \end{subfigure}
  \end{subfigure}
  \caption{Impostor and genuine distributions for ArcFace matcher and multiple datasets. The female impostor distribution ranges over higher similarity scores, indicating a higher FMR, and genuine distribution ranges over lower similarity scores, indicating a higher FNMR.}
  \label{fig:auth_imp}
\end{figure*}

\section{Lower Recognition Accuracy For Females}

The results in Figure \ref{fig:auth_imp} are 
representative of the consensus across many different matchers and datasets as covered in the Related Work. 
Note that the female impostor distribution ranges over higher similarity scores, indicating a higher false match rate (FMR), 
and also that the female genuine distribution ranges over lower similarity scores, indicating a higher false non-match rate (FNMR).

The MORPH dataset \cite{MORPH, MORPH_site} was originally collected to support research in face aging, and has been widely used in that context.
In the last few years, it has also been used in the study of demographic variation in accuracy \cite{KrishnapriyaTTS, Albiero2020_train, Albiero2020_gender, albiero2019does, krishnapria_cvprw_2019}.
MORPH contains mugshot-style images that are nominally frontal pose, neutral expression and acquired with controlled lighting and an 18\% gray background.
We curated the MORPH 3 dataset in order to remove duplicate images, twins, and mislabeled images.
The results in Figure \ref{fig:auth_imp} are for a subset of MORPH 3 that contains:
35,276 images of 8,835 Caucasian males,
10,941 images of 2,798 Caucasian females,
56,245 images of 8,839 African-American males,
and
24,857 images of 5,929 African-American females.

The Notre Dame dataset is a subset of the Face Recognition Grand Challenge dataset \cite{frgc}, made available by the authors of a recent work in this area \cite{Albiero2020_gender}.
It contains frontal images, higher resolution than MORPH, with uniform background, for 261 Caucasian males (14,345 images) and 169 Caucasian females (10,021 images).
Thus the number of subjects is an order-of-magnitude smaller than MORPH, and subjects have more images on average. 
%
%
%
%
%

ArcFace \cite{arcface} is a state-of-the-art deep CNN matcher.
The instance of ArcFace used here corresponds to a set of publicly-available weights \cite{insightface},
trained on the MS1MV2 dataset, which is a publicly-available, ``cleaned'' version of MS1M \cite{ms1_celeb}.
%
%
We estimate that MS1MV2 is about 73\% male / 27\% female in number of subjects
and 67\% male / 33\% female in number of images \cite{Albiero2020_train}.
There should be zero overlap of subjects between 
MS1MV2 and the datasets generating the results in Figure \ref{fig:auth_imp}.
%
%

The lower accuracy for females can be quantified in terms of
the separation between the female and male impostor and genuine distributions.
The cross-gender and same-gender d-prime values are shown in Table \ref{tab:d_prime}. For all three datasets, the female impostor distribution is shifted toward higher similarity scores than the male impostor distribution, 
the female impostor distribution is shifted toward lower similarity scores,
with the result that the impostor-to-genuine separation is lower for females. 

\section{Less ``Pixels On the Face'' In Female Face Images}

\begin{figure*}[t]
  \begin{subfigure}[b]{1\linewidth}
      \begin{subfigure}[b]{0.32\linewidth}
        \centering
          \begin{subfigure}[b]{0.35\columnwidth}
            \centering
            \includegraphics[width=\linewidth]{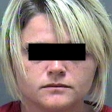}
          \end{subfigure}
          \begin{subfigure}[b]{0.35\columnwidth}
            \centering
            \includegraphics[width=\linewidth]{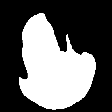}
          \end{subfigure}
          \begin{subfigure}[b]{0.35\columnwidth}
            \centering
            \includegraphics[width=\linewidth]{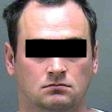}
          \end{subfigure}
          \begin{subfigure}[b]{0.35\columnwidth}
            \centering
            \includegraphics[width=\linewidth]{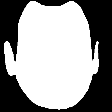}
          \end{subfigure}
          \caption{MORPH Caucasian}
      \end{subfigure}
      \hfill
      \begin{subfigure}[b]{0.32\linewidth}
        \centering
          \begin{subfigure}[b]{0.35\columnwidth}
            \centering
            \includegraphics[width=\linewidth]{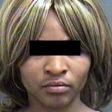}
          \end{subfigure}
          \begin{subfigure}[b]{0.35\columnwidth}
            \centering
            \includegraphics[width=\linewidth]{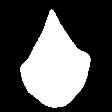}
          \end{subfigure}
          \begin{subfigure}[b]{0.35\columnwidth}
            \centering
            \includegraphics[width=\linewidth]{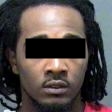}
          \end{subfigure}
          \begin{subfigure}[b]{0.35\columnwidth}
            \centering
            \includegraphics[width=\linewidth]{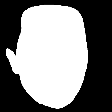}
          \end{subfigure}
          \caption{MORPH African-American}
      \end{subfigure}
      \hfill
        \begin{subfigure}[b]{0.32\linewidth}
          \centering
          \begin{subfigure}[b]{0.35\columnwidth}
            \centering
            \includegraphics[width=\linewidth]{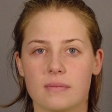}
          \end{subfigure}
          \begin{subfigure}[b]{0.35\columnwidth}
            \centering
            \includegraphics[width=\linewidth]{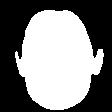}
          \end{subfigure}
          \begin{subfigure}[b]{0.35\columnwidth}
            \centering
            \includegraphics[width=\linewidth]{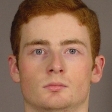}
          \end{subfigure}
          \begin{subfigure}[b]{0.35\columnwidth}
            \centering
            \includegraphics[width=\linewidth]{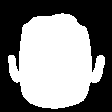}
          \end{subfigure}
          \caption{Notre Dame Caucasian}
      \end{subfigure}
  \end{subfigure}
  \caption{Example images and their face/non-face masks based on BiSeNet segmentation (eye regions of MORPH images blacked out for this figure as a privacy consideration).}
  \label{fig:mean_faces}
\end{figure*}
\begin{table}[t]
    \centering
    \ra{1.2}
    \scriptsize
   \begin{tabular}{l|p{1.8cm}|p{1.8cm}|p{1.8cm}|p{1.8cm}}
     & \multicolumn{2}{c|}{\textbf{Cross Gender d-prime}} & \multicolumn{2}{c}{\textbf{Same Gender Genuine-Impostor d-prime}} \\
    \textbf{Dataset} & \textbf{Genuine} & \textbf{Impostor} & \textbf{Male} & \textbf{Female} \\ \hline
    \textbf{MORPH Caucasian} & 0.197 & 0.266 & 9.19 & 8.38 \\
    \textbf{MORPH African-American} & 0.314 & 0.458 & 9.97 & 8.32 \\
    \textbf{Notre Dame} & 0.087 & 0.271 & 11.58 & 10.54
    \end{tabular}
    \vspace{1.5em}
    \caption{Cross gender d-prime (lower is better, more similar) and same gender d-prime (higher is better, more separation) MORPH and Notre Dame datasets.}
 \label{tab:d_prime}
\end{table}
\begin{figure*}[t]
  \begin{subfigure}[b]{1\linewidth}
    \begin{subfigure}[b]{0.32\linewidth}
      \begin{subfigure}[b]{0.48\columnwidth}
        \centering
          \begin{subfigure}[b]{1\columnwidth}
            \centering
            \includegraphics[width=\linewidth]{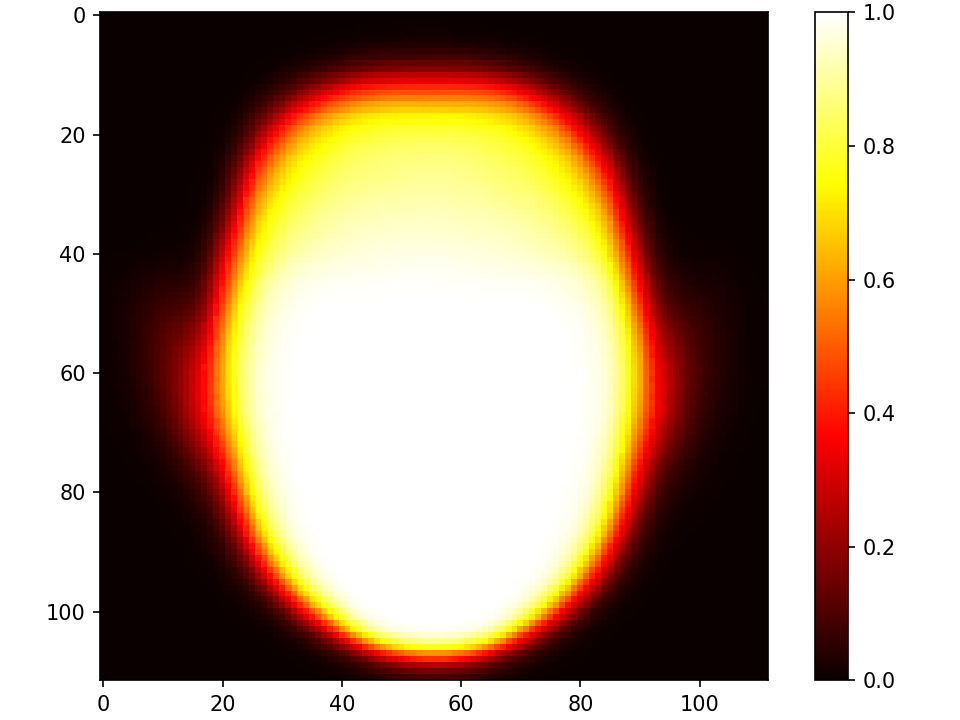}
          \end{subfigure}
          \begin{subfigure}[b]{1\columnwidth}
            \centering
            \includegraphics[width=\linewidth]{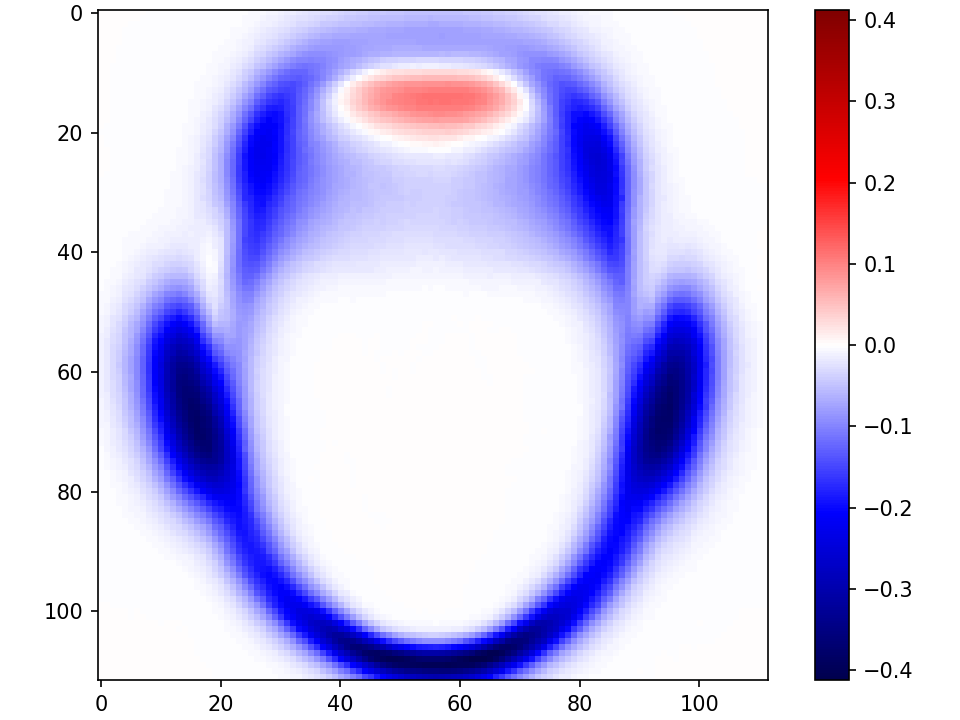}
          \end{subfigure}
          \begin{subfigure}[b]{1\columnwidth}
            \centering
            \includegraphics[width=\linewidth]{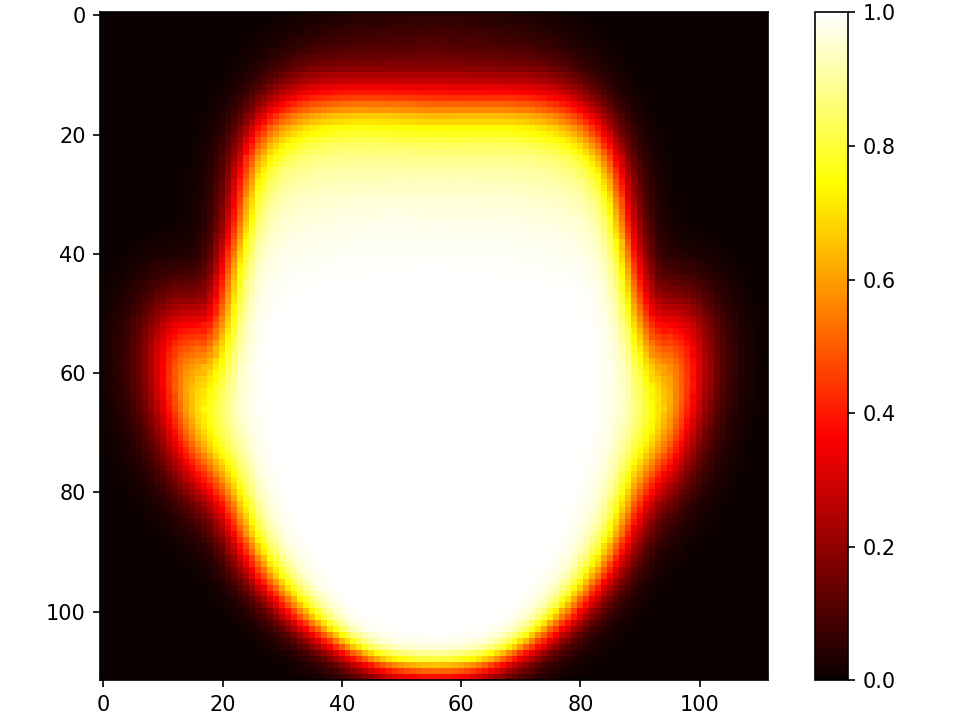}
          \end{subfigure}
      \end{subfigure}
      \begin{subfigure}[b]{0.48\columnwidth}
        \centering
          \begin{subfigure}[b]{1\columnwidth}
            \centering
            \includegraphics[width=\linewidth]{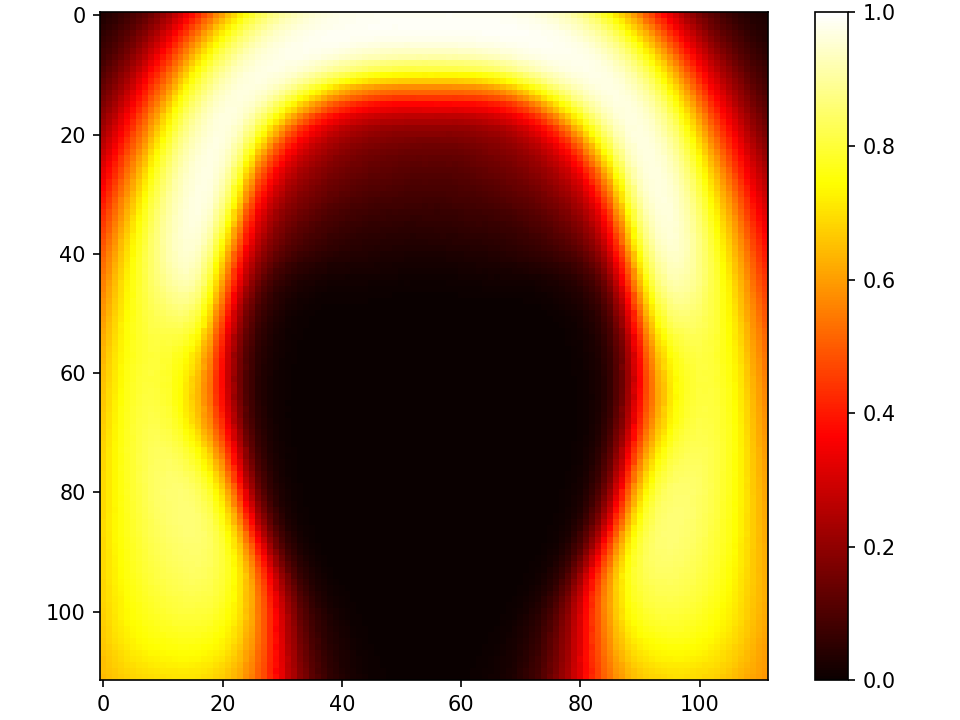}
          \end{subfigure}
          \begin{subfigure}[b]{1\columnwidth}
            \centering
            \includegraphics[width=\linewidth]{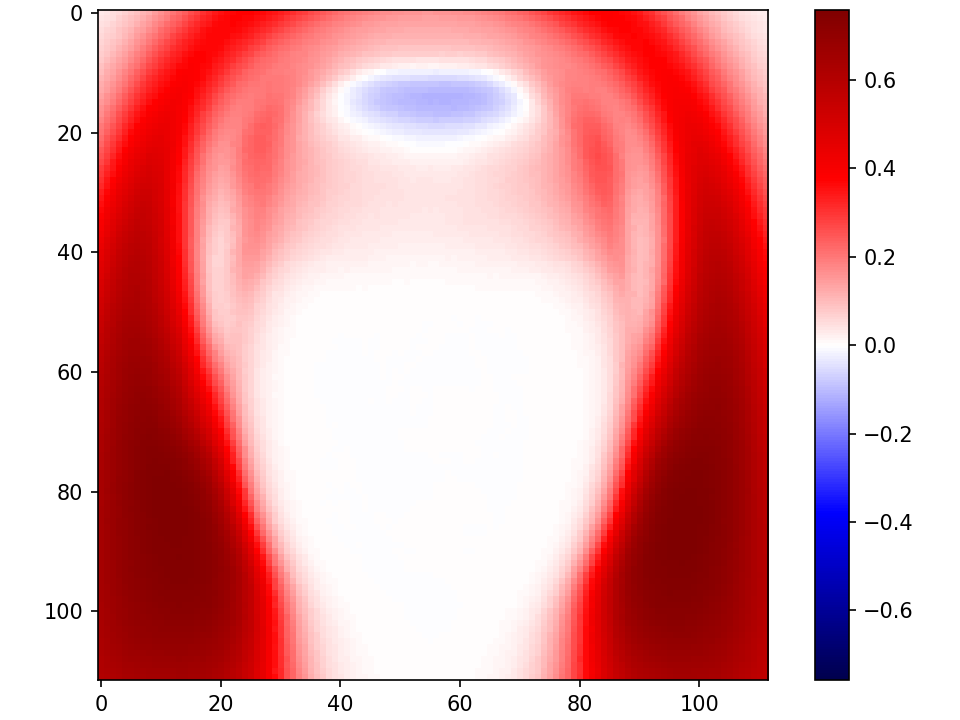}
          \end{subfigure}
          \begin{subfigure}[b]{1\columnwidth}
            \centering
            \includegraphics[width=\linewidth]{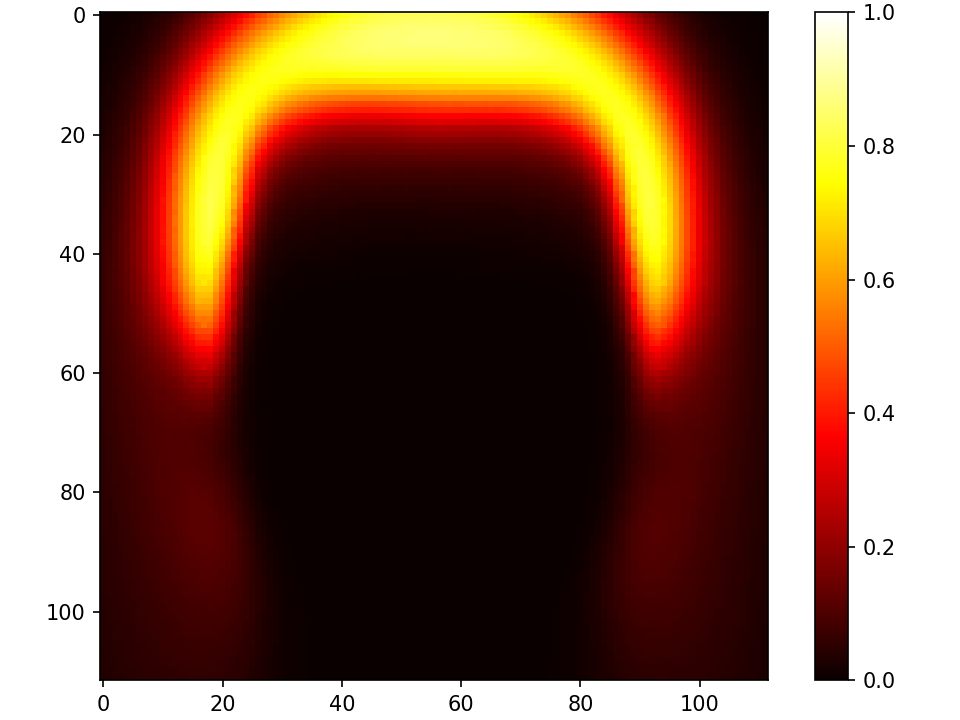}
          \end{subfigure}
        \end{subfigure}
          \caption{MORPH Caucasian}
      \end{subfigure}
      \hfill 
      \begin{subfigure}[b]{0.32\linewidth}
      \begin{subfigure}[b]{0.48\columnwidth}
        \centering
          \begin{subfigure}[b]{1\columnwidth}
            \centering
            \includegraphics[width=\linewidth]{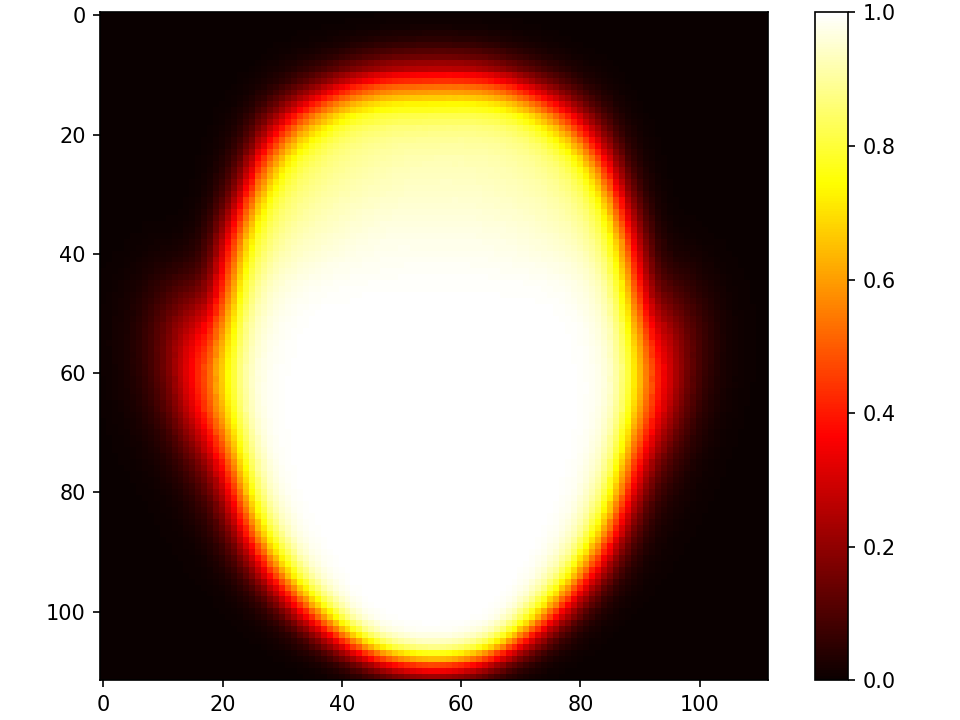}
          \end{subfigure}
          \begin{subfigure}[b]{1\columnwidth}
            \centering
            \includegraphics[width=\linewidth]{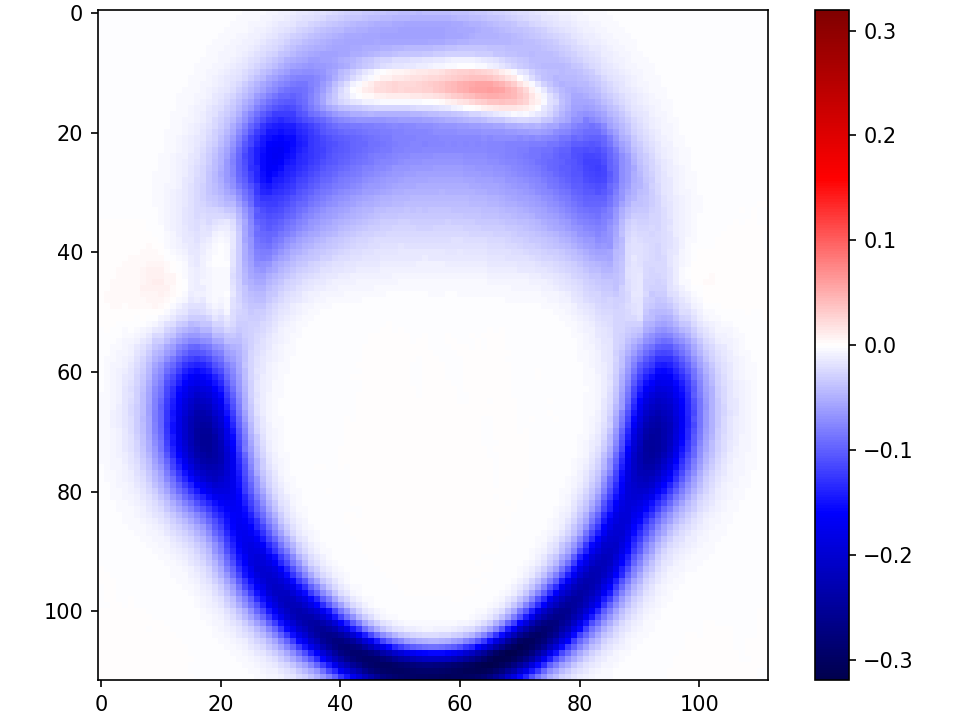}
          \end{subfigure}
          \begin{subfigure}[b]{1\columnwidth}
            \centering
            \includegraphics[width=\linewidth]{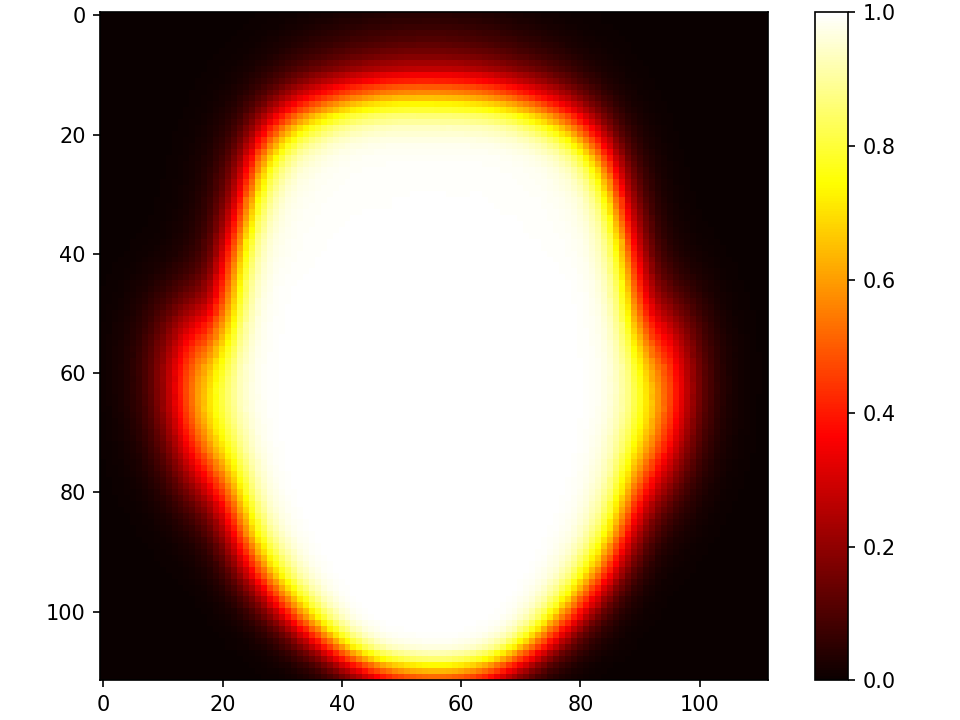}
          \end{subfigure}
      \end{subfigure}
          \begin{subfigure}[b]{0.48\columnwidth}
        \centering
          \begin{subfigure}[b]{1\columnwidth}
            \centering
            \includegraphics[width=\linewidth]{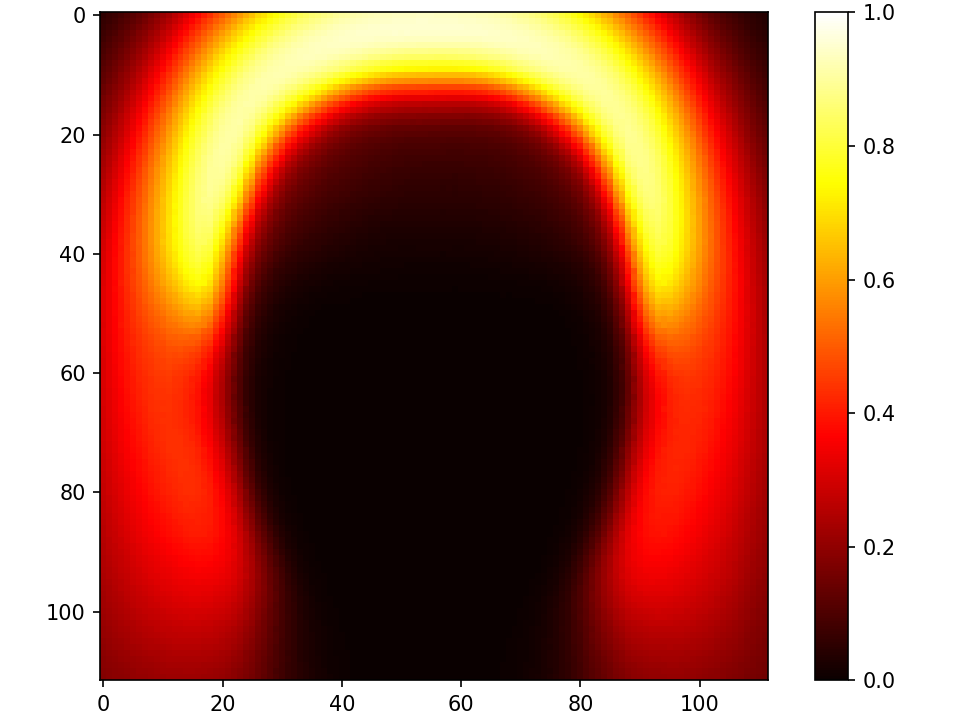}
          \end{subfigure}
          \begin{subfigure}[b]{1\columnwidth}
            \centering
            \includegraphics[width=\linewidth]{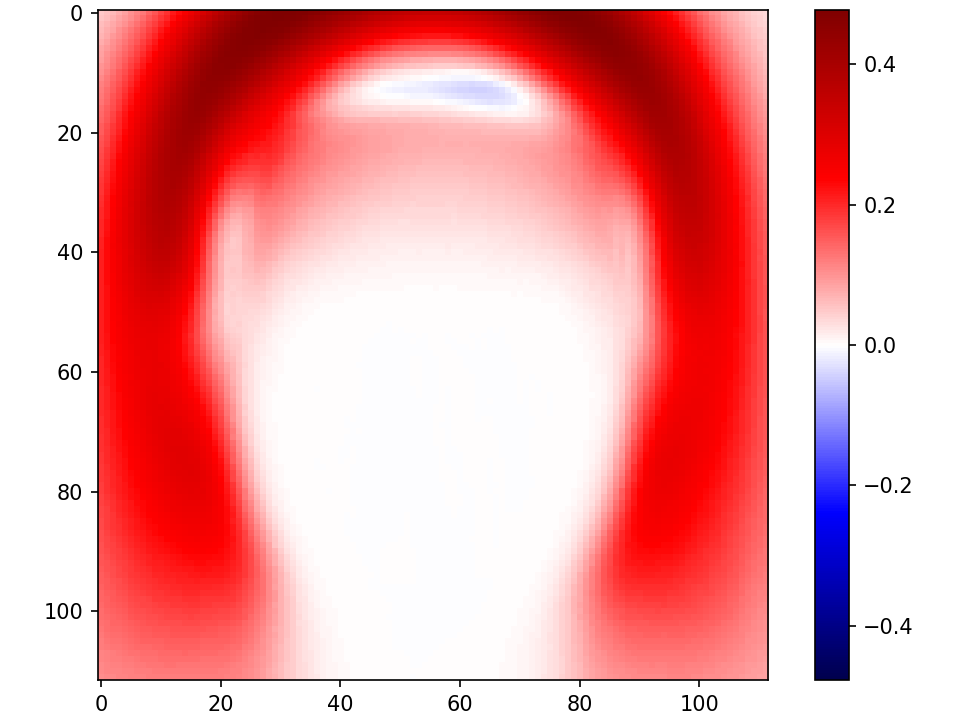}
          \end{subfigure}
          \begin{subfigure}[b]{1\columnwidth}
            \centering
            \includegraphics[width=\linewidth]{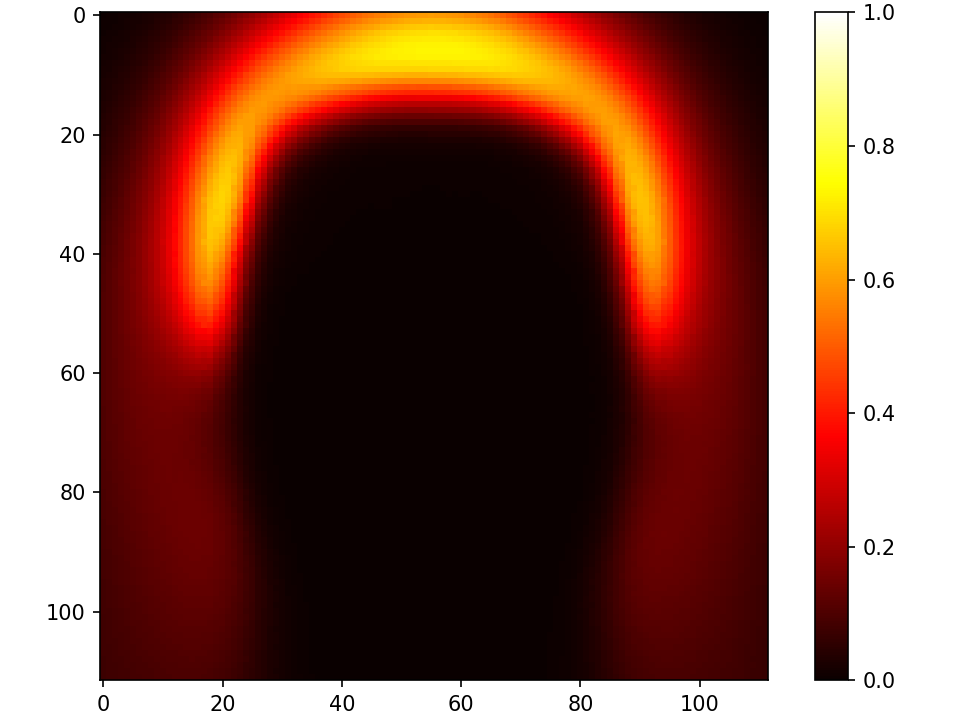}
          \end{subfigure}
        \end{subfigure}
          \caption{MORPH African-American}
      \end{subfigure}
      \hfill 
      \begin{subfigure}[b]{0.32\linewidth}
      \begin{subfigure}[b]{0.48\columnwidth}
        \centering
          \begin{subfigure}[b]{1\columnwidth}
            \centering
            \includegraphics[width=\linewidth]{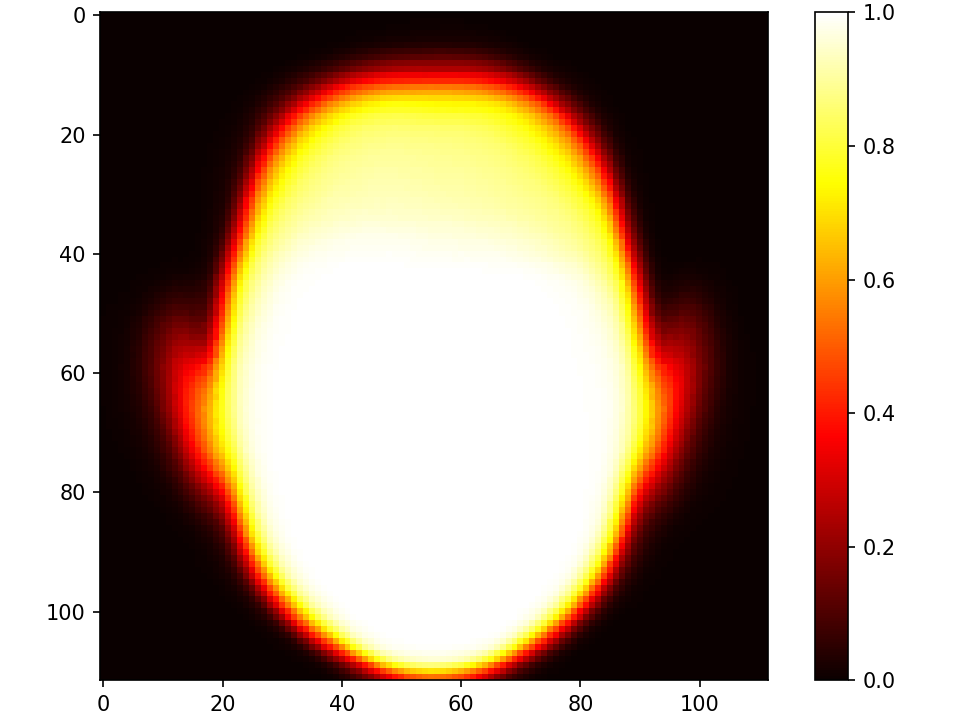}
          \end{subfigure}
          \begin{subfigure}[b]{1\columnwidth}
            \centering
            \includegraphics[width=\linewidth]{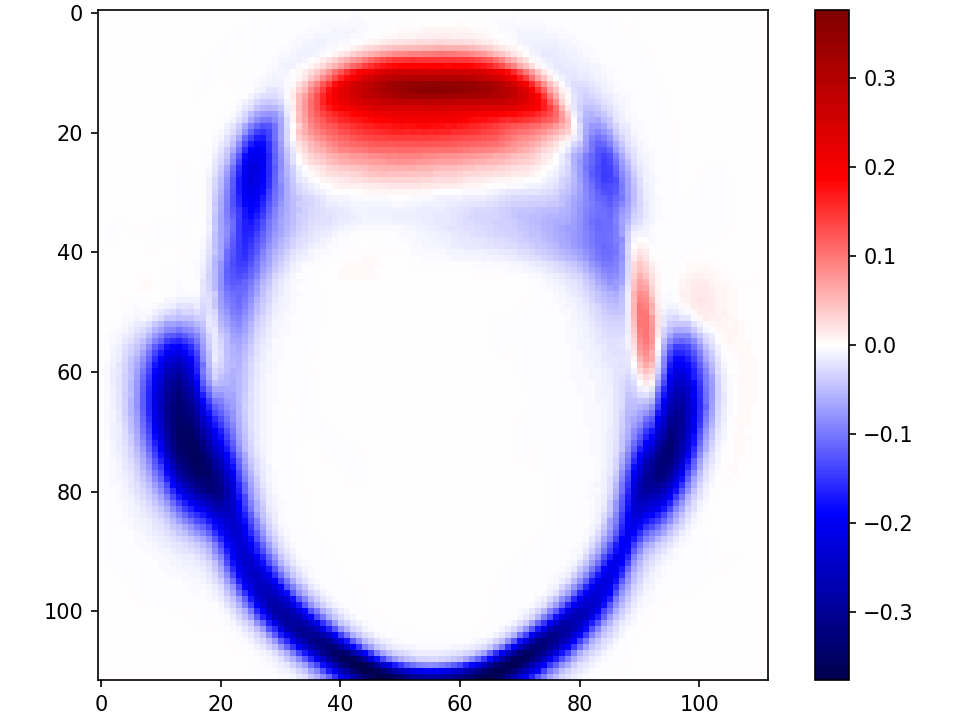}
          \end{subfigure}
          \begin{subfigure}[b]{1\columnwidth}
            \centering
            \includegraphics[width=\linewidth]{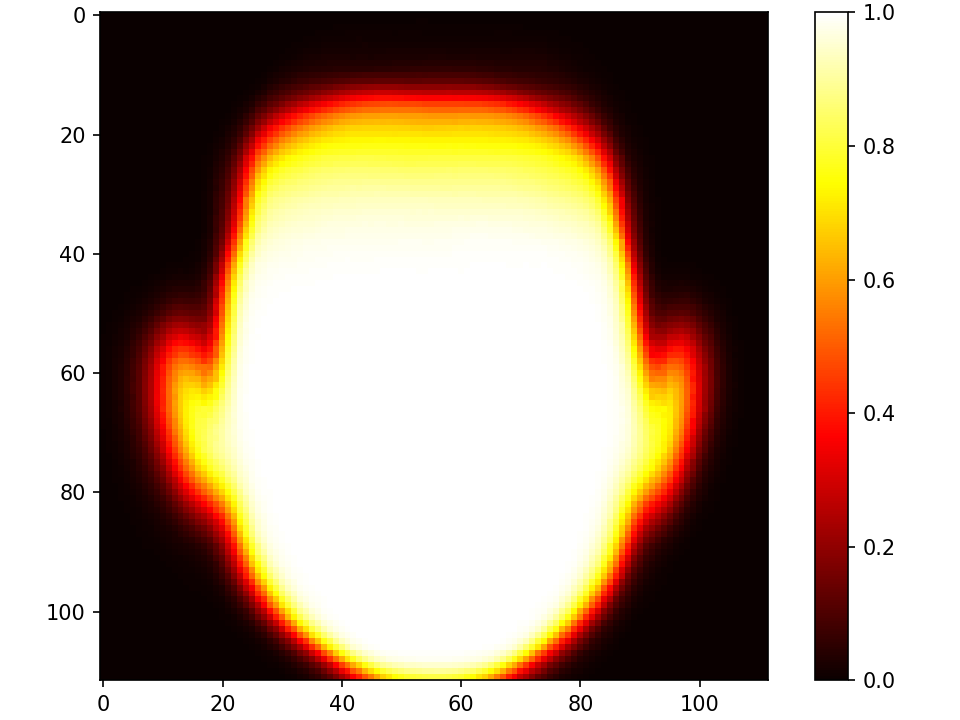}
          \end{subfigure}
      \end{subfigure}
      \begin{subfigure}[b]{0.48\columnwidth}
        \centering
          \begin{subfigure}[b]{1\columnwidth}
            \centering
            \includegraphics[width=\linewidth]{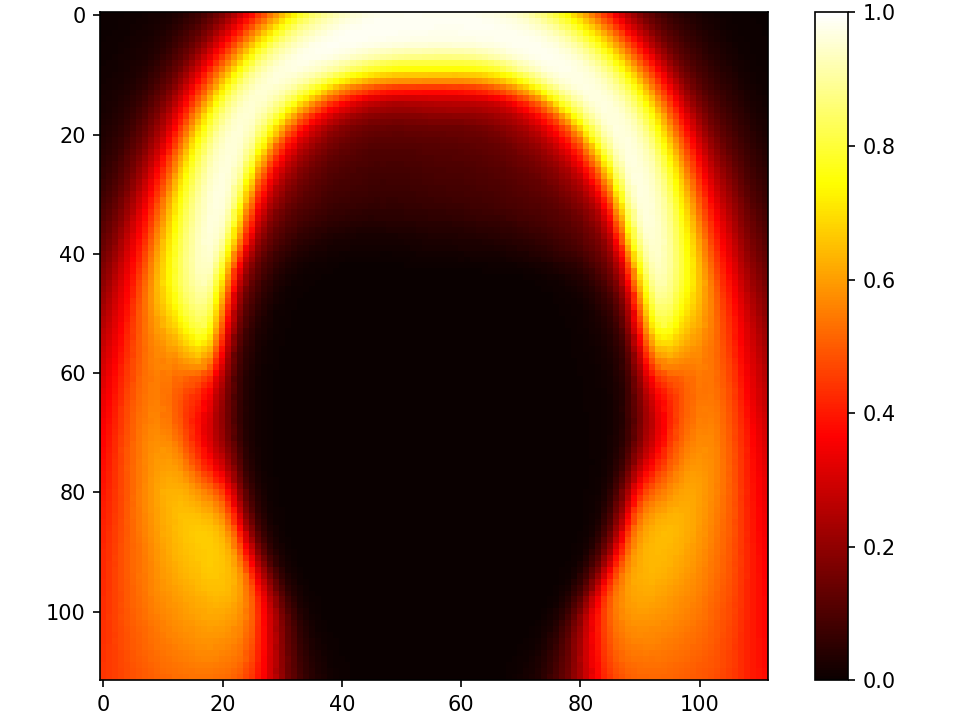}
          \end{subfigure}
          \begin{subfigure}[b]{1\columnwidth}
            \centering
            \includegraphics[width=\linewidth]{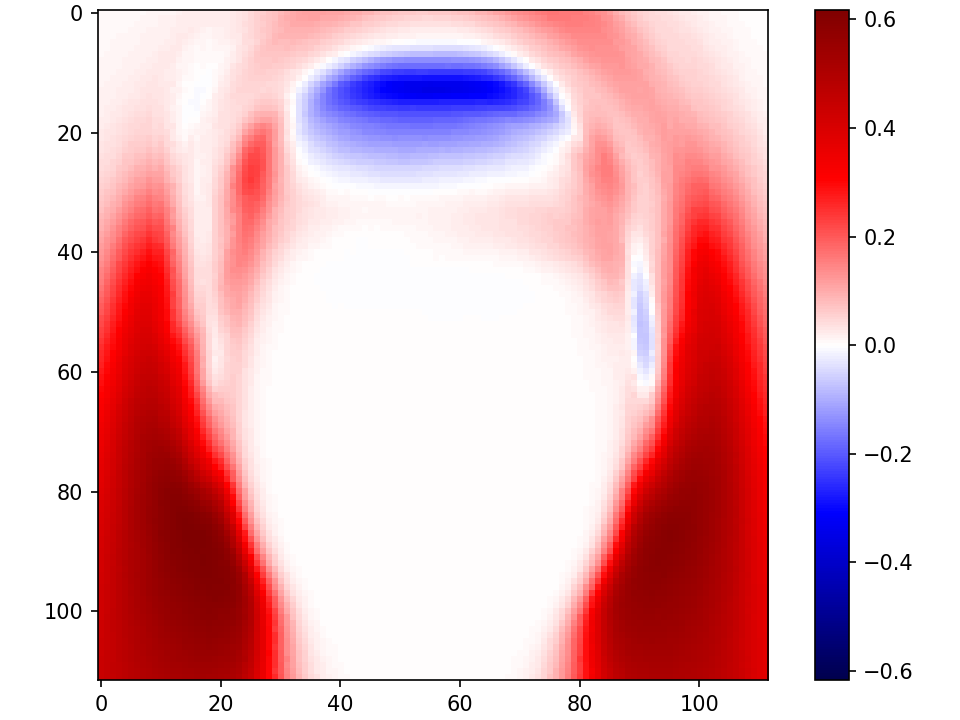}
          \end{subfigure}
          \begin{subfigure}[b]{1\columnwidth}
            \centering
            \includegraphics[width=\linewidth]{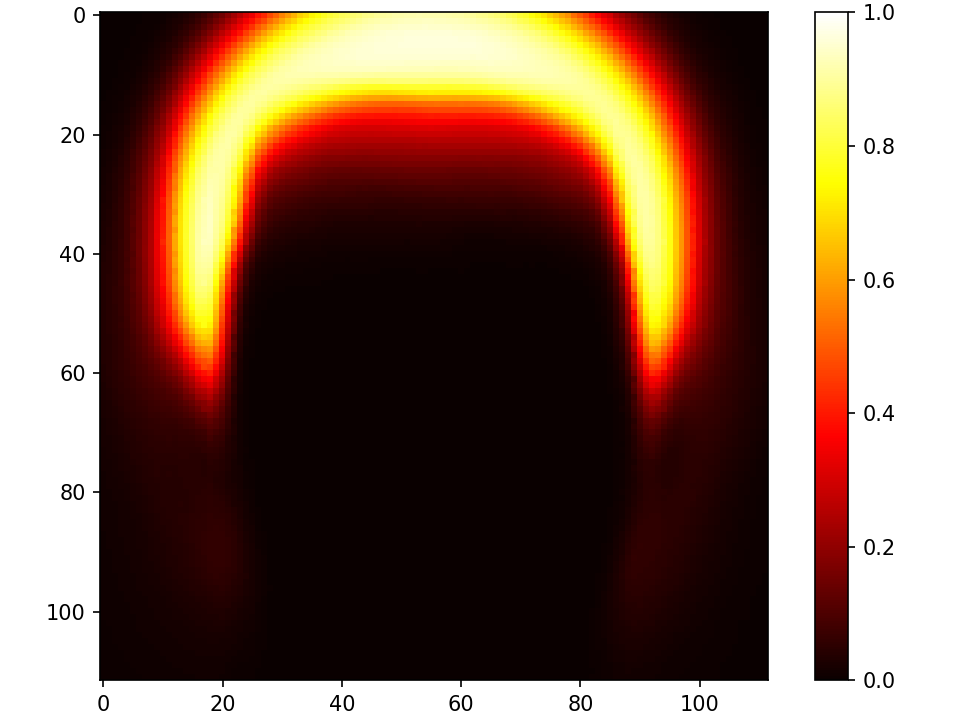}
          \end{subfigure}
         \end{subfigure}
          \caption{Notre Dame Caucasian}
      \end{subfigure}
  \end{subfigure}
  \caption{Heatmaps representing frequency of pixels being labelled as ``face'' on left and as hair on right. Female heatmap is on top, male heatmap is on bottom, and male-female difference is on middle.}
  \label{fig:heatmap}
\end{figure*}

This section considers the face images that are input to the deep CNN,
the fraction of the image that represents ``pixels on the face'', and how this differs between female and male.
The faces in the MORPH and Notre Dame images are detected and aligned using RetinaFace \cite{retinaface}. 
This results in the 112x112 image with face centered in standard location that is the input to the deep CNN.
Examples are shown in Figure \ref{fig:mean_faces}.
%
%
To obtain a binary mask that indicates which pixels represent face and which do not, we use
the modified Bilateral Segmentation Network (``BiSeNet'') to segment the faces 
(\cite{bisenet}, implementation at \cite{bisenet_github}).
A pre-trained version of BiSeNet segments face images into regions.
For our purposes, ``face'' is the union of BiSeNet regions 1 to 13, corresponding to skin, eyebrows, eyes, ears, nose and mouth. 
%
%
BiSeNet regions corresponding to neck, clothes, hair and hat are excluded.
Examples of the face/non-face masks are shown in Figure \ref{fig:mean_faces}.
These masks are used to compute heatmaps for face being visible at each pixel, for female and male images. 

The value of each heatmap pixel is from 0 to 1, reflecting the fraction of the female (male) face images for which that pixel is labelled as face.
Figure \ref{fig:heatmap} shows the female and male heatmaps and the difference between them.
The chin in the male heatmap extends slightly further toward the bottom.
This reflects the ``biology'' of females and males having, on average, different head size / shape.
Another difference is that the ear region and the sides of the face are less prominent in the female heatmap.
This reflects gender-associated hairstyles, where female hairstyles more frequently occlude the ears and part of the sides of the face.
The difference heatmaps summarize all of this, with blue representing pixels that more frequently labeled face for males than for females, and red representing the opposite.
The heatmaps make it clear that, {\it on average, female face images contain fewer ``pixels on the face'' than male face images, and that gendered hairstyles are a major underlying cause.}

%
%

\begin{figure*}[t]
  \begin{subfigure}[b]{1\linewidth}
      \begin{subfigure}[b]{0.32\linewidth}
        \centering
          \includegraphics[width=\linewidth]{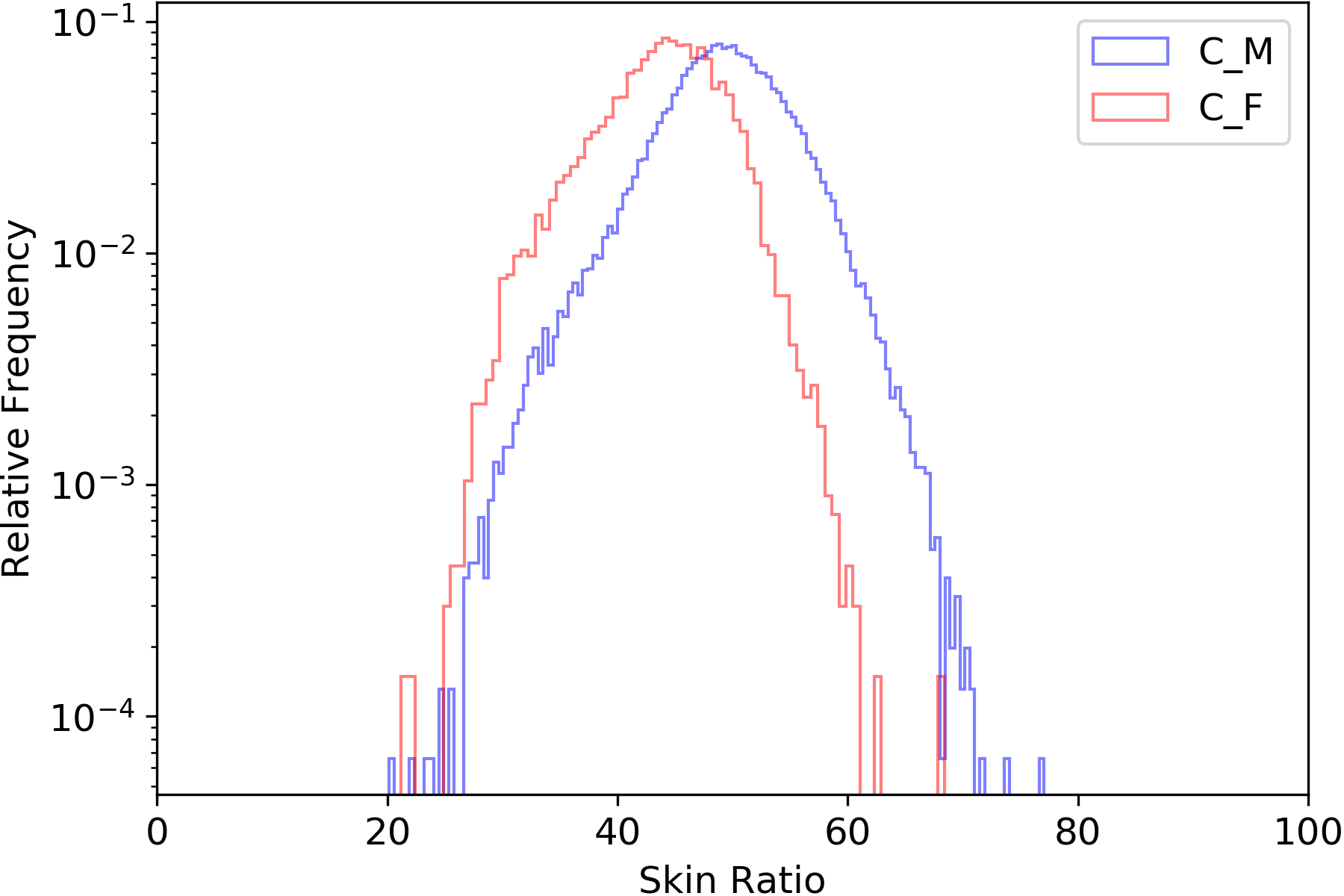}
          \caption{MORPH Caucasian}
          \vspace{-0.5em}
      \end{subfigure}
      \hfill 
      \begin{subfigure}[b]{0.32\linewidth}
        \centering
          \includegraphics[width=\linewidth]{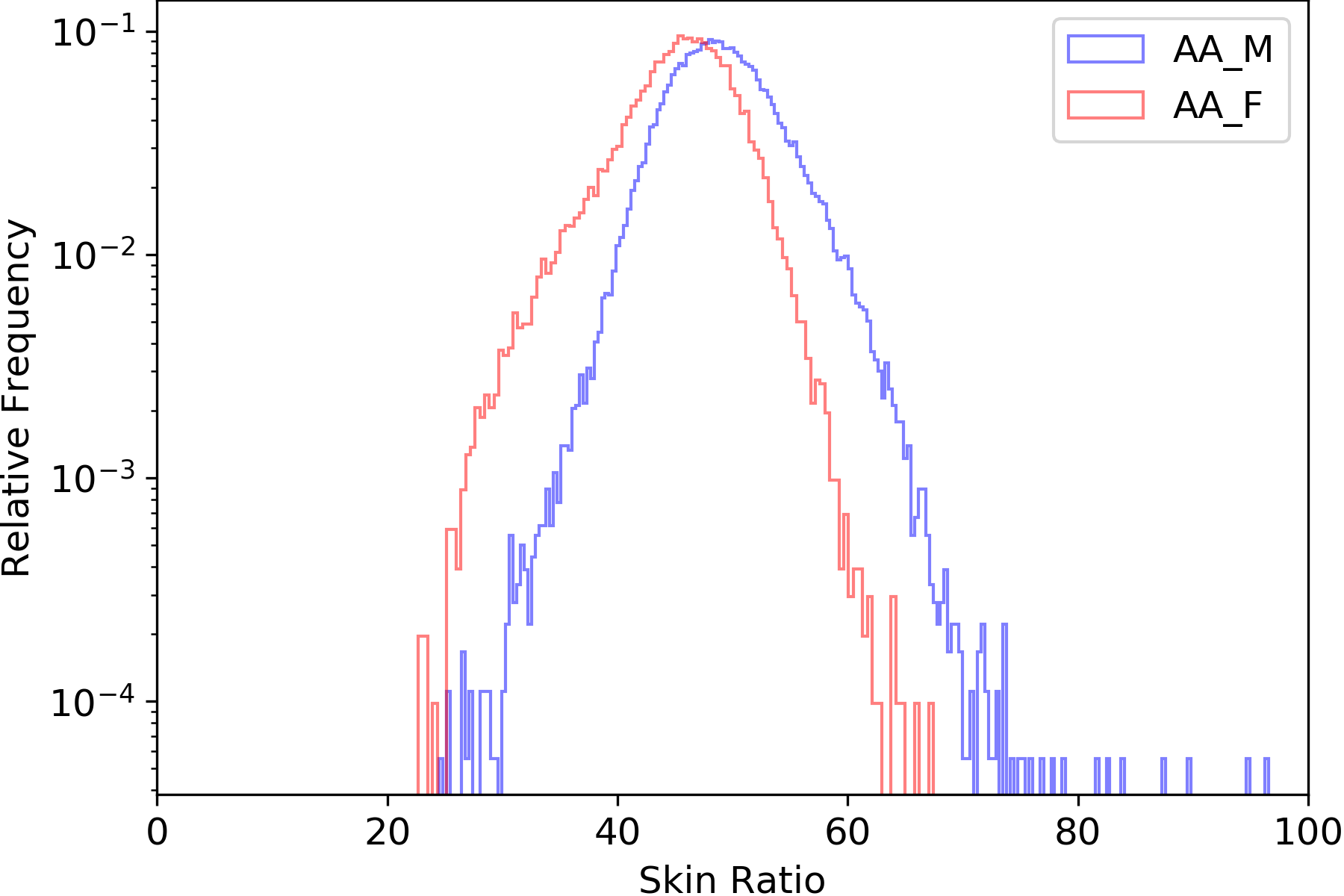}
          \caption{MORPH African-American}
          \vspace{-0.5em}
      \end{subfigure}
      \hfill 
      \begin{subfigure}[b]{0.32\linewidth}
        \centering
          \includegraphics[width=\linewidth]{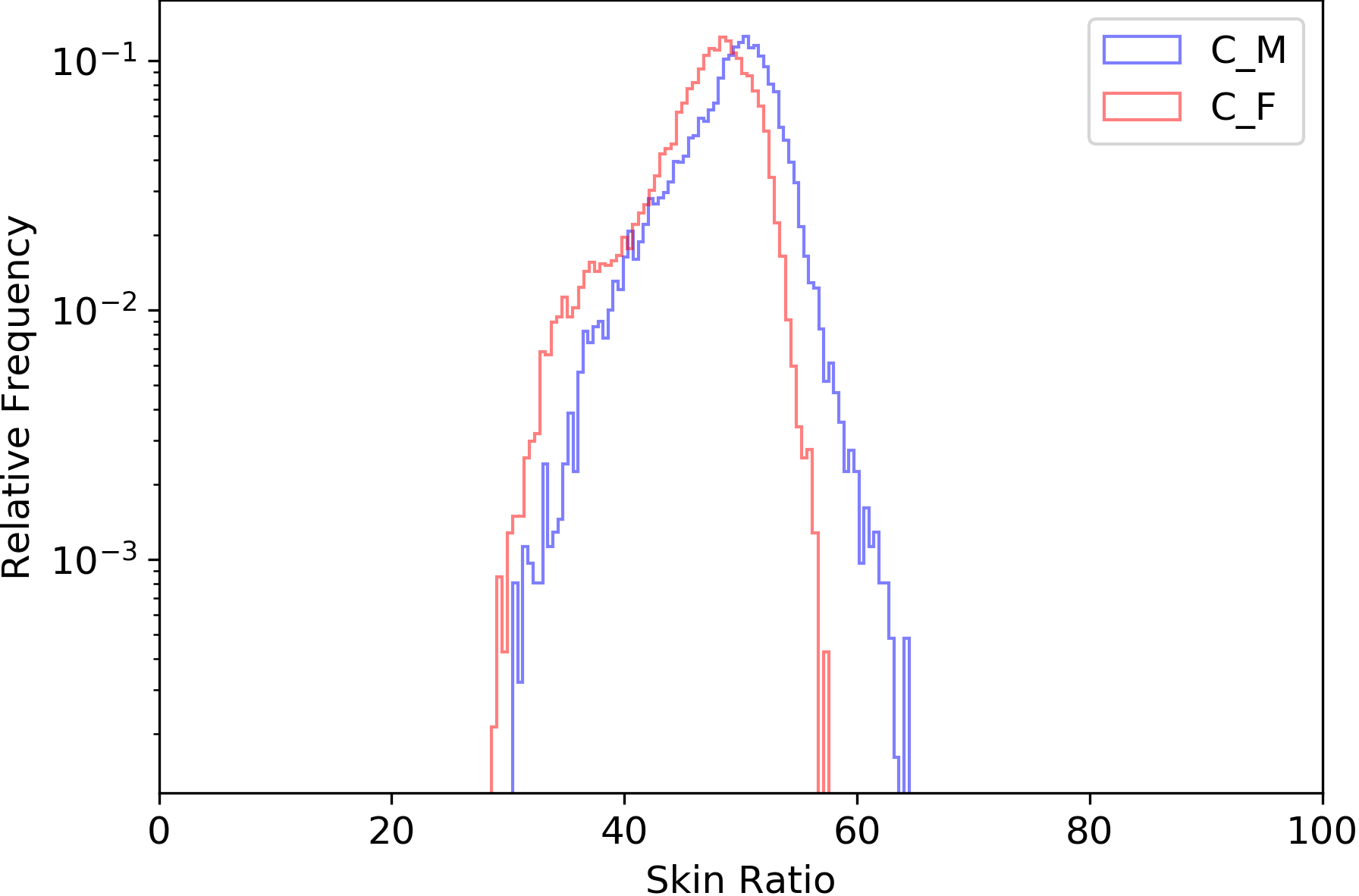}
          \caption{Notre Dame Caucasian}
          \vspace{-0.5em}
      \end{subfigure}
  \end{subfigure}
  \caption{Comparison of female / male distributions of percent of image labelled face.}
  \vspace{-0.5em}
  \label{fig:skin_dist}
\end{figure*}

A different view of this information is the distribution of the fraction of the image that represents the face.
The ``\% face'' distributions compared in Figure \ref{fig:skin_dist} show that the proportion of female face images is larger in the range of approximately 25\% to 45\%, and the proportion of male face images is larger in the range of about 45\% to 70\%.
Again it is clear that, 
on average, the female face images contain less information about the face.

Note the correlation of the difference in the \% face distributions in Figure \ref{fig:skin_dist} and the difference in the genuine distributions in Figure \ref{fig:auth_imp}.
The most pronounced difference in the \% face distributions is for MORPH African-American, which also has the most pronounced difference in genuine distributions.
The most similar \% face distributions are for Notre Dame Caucasian, which also has the most similar genuine distributions.

\section{Equal Face Info Flips Genuine Distribution Difference}

The previous section shows that the observation that face recognition accuracy is lower for females is based on 
images in which the female face images contain less information about the face.
Given this, it is natural to ask what happens if the dataset is controlled to have equal information in female and male face images.

\begin{figure*}[t]

  \begin{subfigure}[b]{0.32\linewidth}
      \begin{subfigure}[b]{1\columnwidth}
        \centering
          \includegraphics[width=\columnwidth]{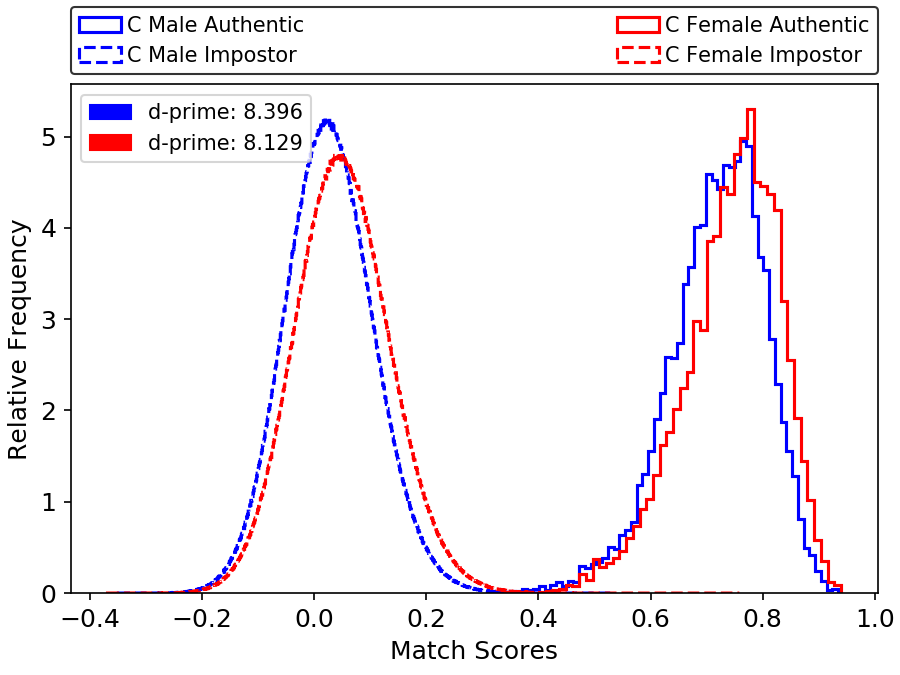}
      \end{subfigure}
      \hfill
      \begin{subfigure}[b]{1\columnwidth}
          \centering
          \begin{subfigure}[b]{0.48\columnwidth}
            \centering
              \includegraphics[width=1\columnwidth]{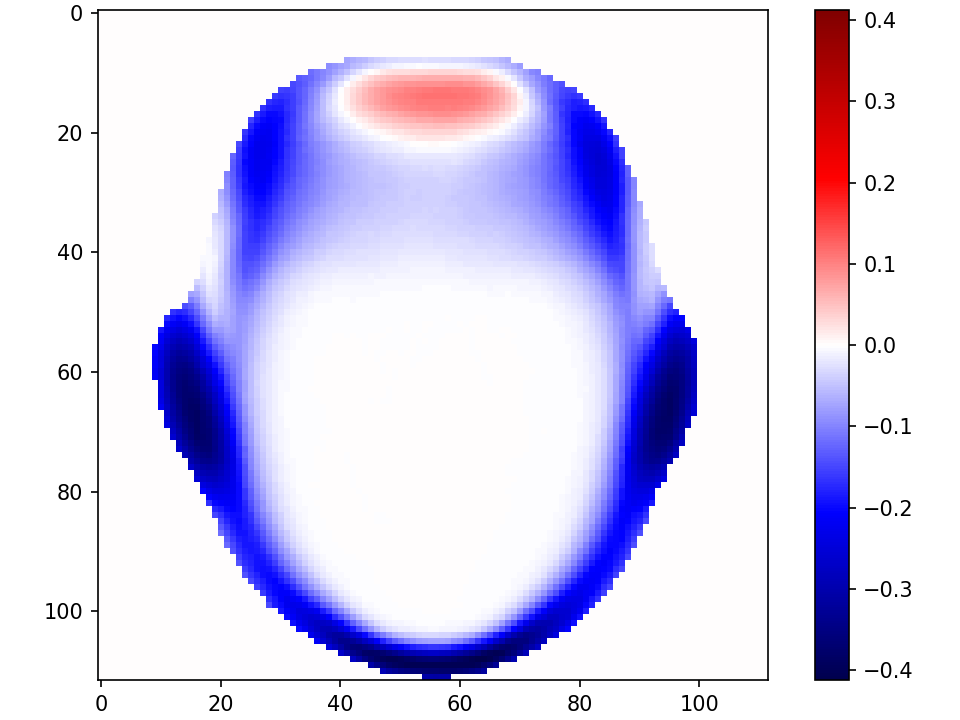}
          \end{subfigure}
          \begin{subfigure}[b]{0.48\columnwidth}
            \centering
              \includegraphics[width=1\columnwidth]{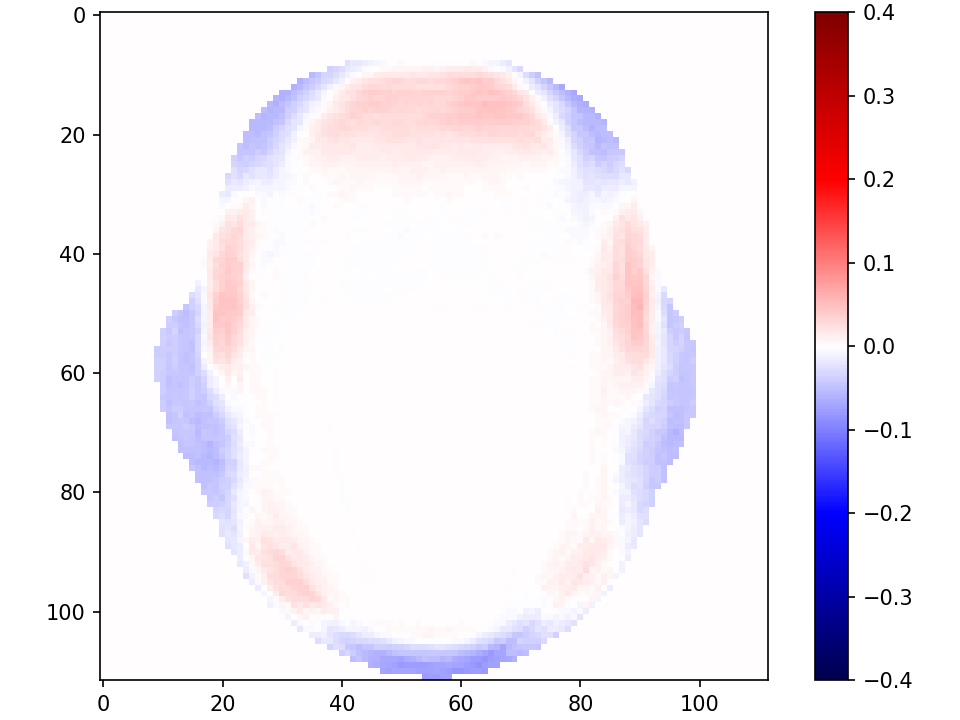}
          \end{subfigure}
      \end{subfigure}
      \caption{MORPH Caucasian}
  \end{subfigure}
  \hfill
  \begin{subfigure}[b]{0.32\linewidth}
      \begin{subfigure}[b]{1\columnwidth}
        \centering
          \includegraphics[width=\columnwidth]{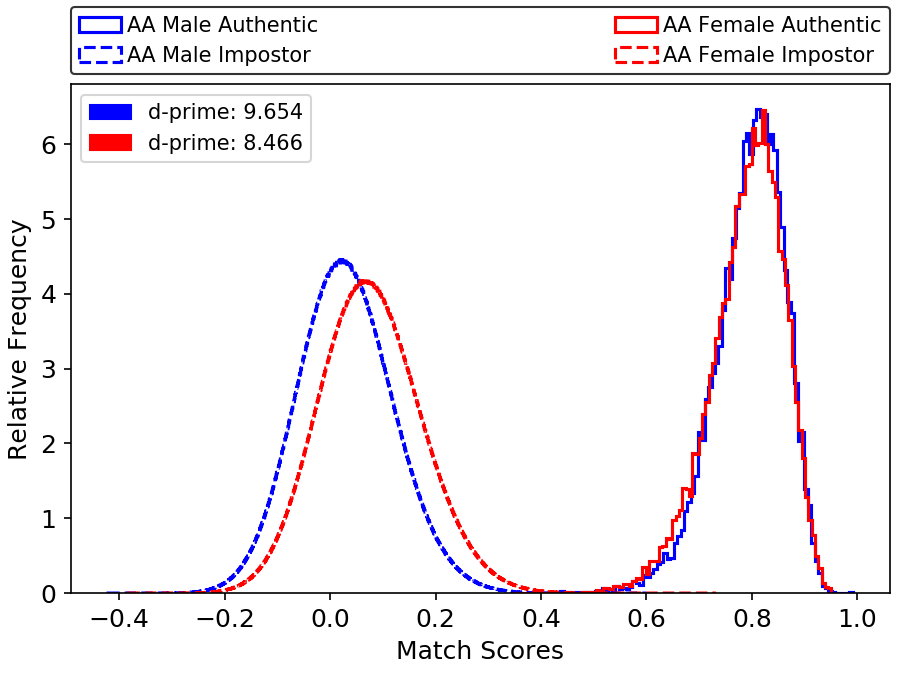}
      \end{subfigure}
      \hfill
      \begin{subfigure}[b]{1\columnwidth}
          \centering
          \begin{subfigure}[b]{0.48\columnwidth}
            \centering
              \includegraphics[width=1\columnwidth]{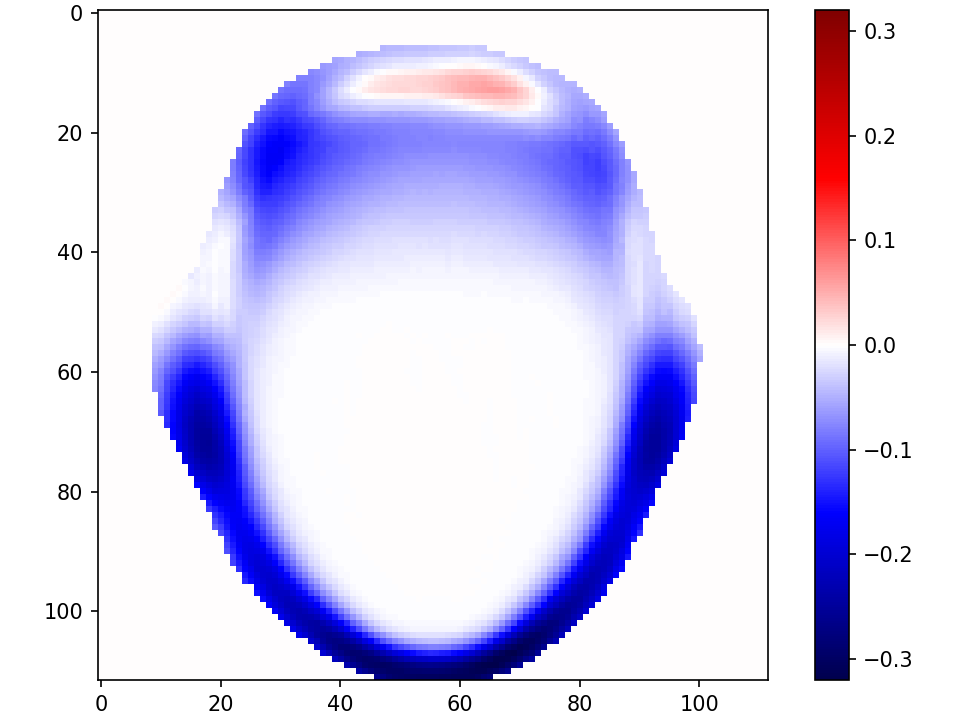}
          \end{subfigure}
          \begin{subfigure}[b]{0.48\columnwidth}
            \centering
              \includegraphics[width=1\columnwidth]{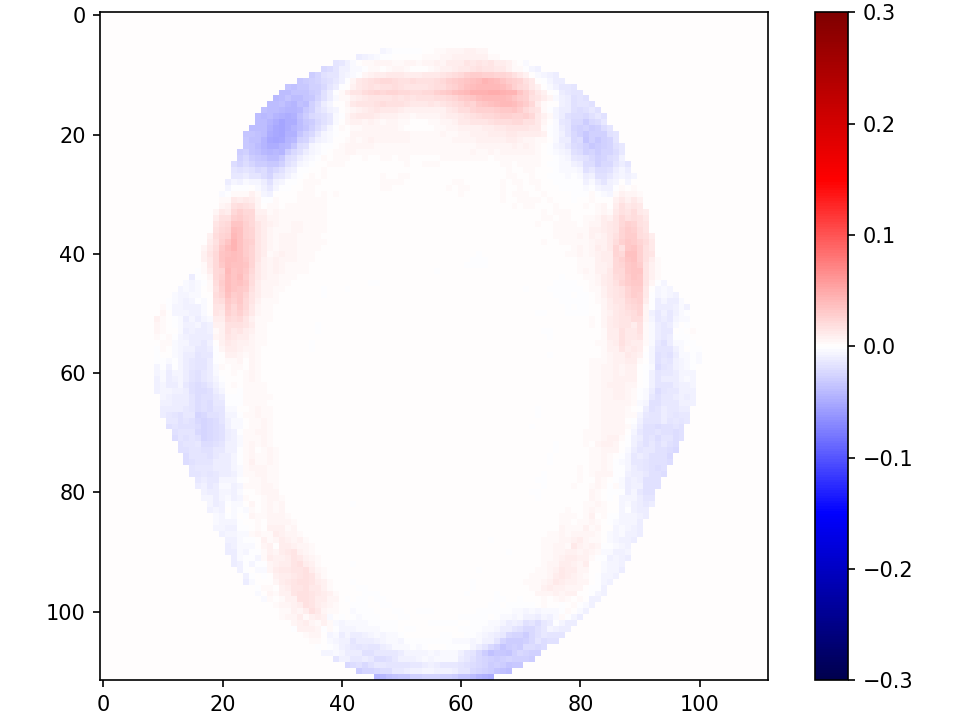}
          \end{subfigure}
      \end{subfigure}
      \caption{MORPH African-American}
  \end{subfigure}
  \hfill
  \begin{subfigure}[b]{0.32\linewidth}
      \begin{subfigure}[b]{1\columnwidth}
        \centering
          \includegraphics[width=\columnwidth]{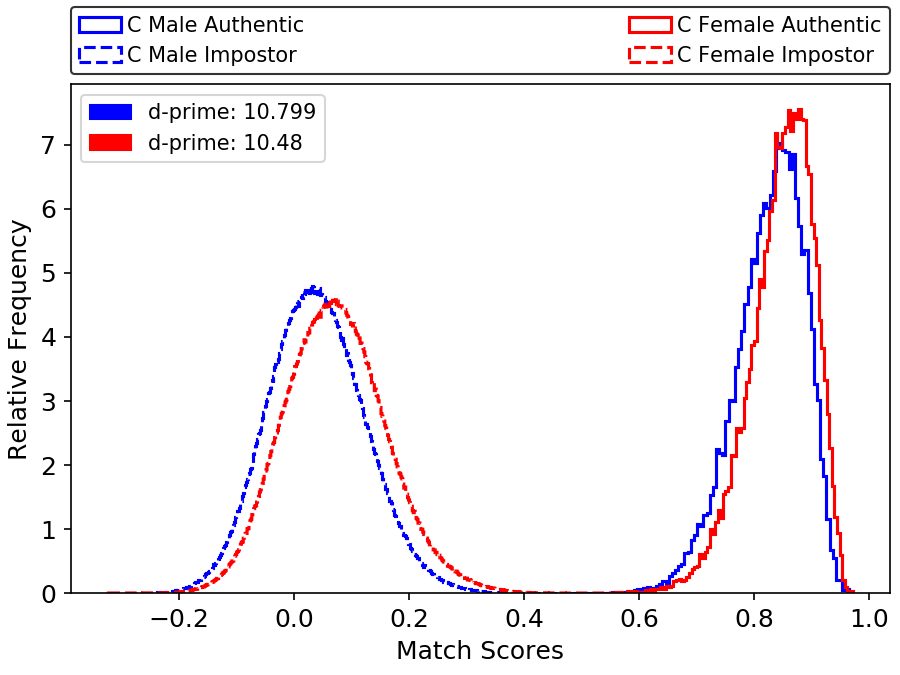}
      \end{subfigure}
      \hfill
      \begin{subfigure}[b]{1\columnwidth}
          \centering
          \begin{subfigure}[b]{0.48\columnwidth}
            \centering
              \includegraphics[width=1\columnwidth]{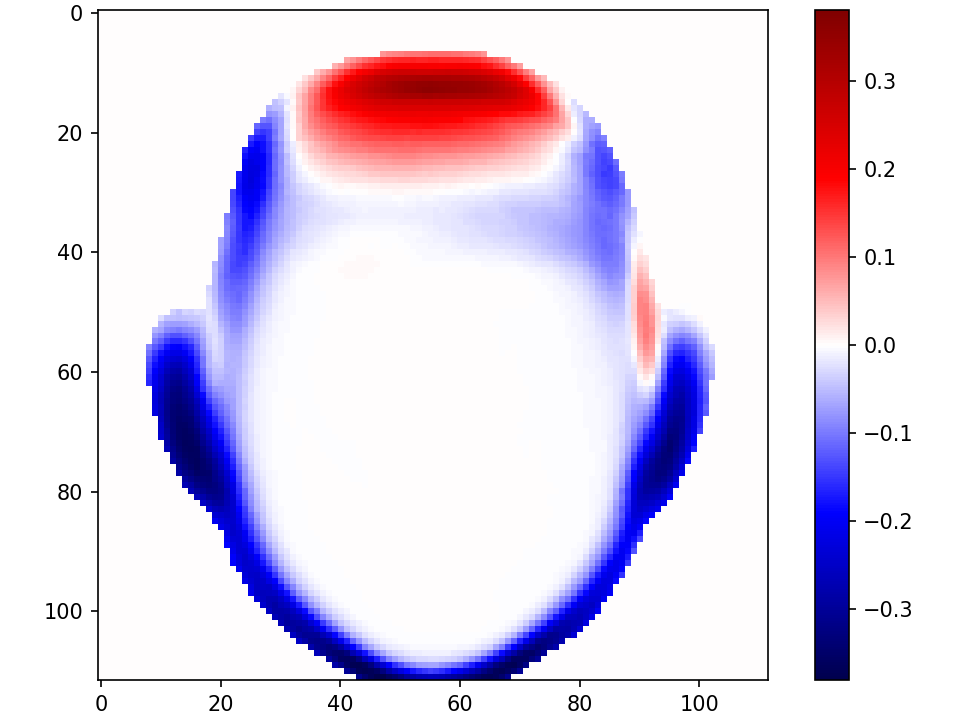}
          \end{subfigure}
          \begin{subfigure}[b]{0.48\columnwidth}
            \centering
              \includegraphics[width=1\columnwidth]{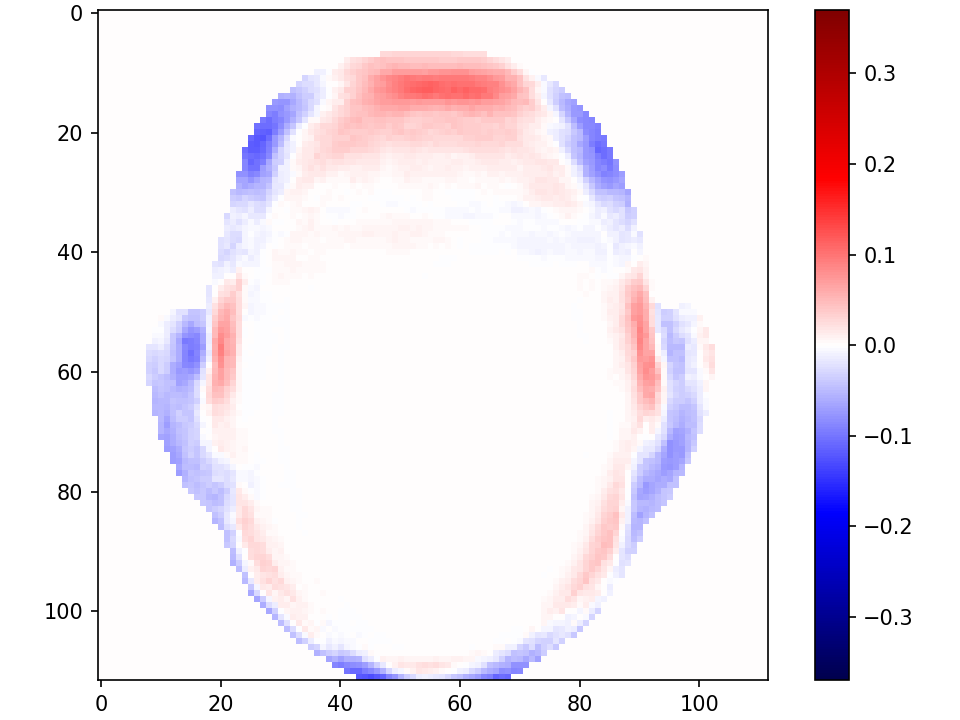}
          \end{subfigure}
      \end{subfigure}
      \caption{Notre Dame}
  \end{subfigure}
  \caption{Impostor and genuine distributions for info-equalized female and male image sets. The original image set is first masked at 10\% level from female heatmap, giving the difference heatmap on lower left. A subset of the male images is selected using the intersection-over-union (IoU) to balance with the female images, giving the difference heatmap on lower right. This information-equalized set of images results in the genuine and impostor distributions in the upper row.}
  \label{fig:equalized_images}
\end{figure*}

We created a version of the dataset designed to minimize the difference in information between the female and male face images.
First, we masked the area outside of the 10\% level in the female heatmap, in all female and male images.
Because the female face region is generally a subset of the male,
this establishes the same maximum set of pixels containing face information in all images, and preserves a large majority of the available face information.
However, this still
leaves a bias of greater information content for male images because, as the comparison heatmap shows, any given pixel is generally labeled as face for a higher fraction of the male images than the female images.
Therefore, as a second step, for each female image, we selected the male image that had the maximum intersection-over-union (IoU) of pixels labeled as face in the two images.
The difference heatmap for this nearly info-equalized set of images,
and the resulting impostor and genuine distributions,
are shown in Figure \ref{fig:equalized_images}.
These results are for a set of 4,730 Caucasian female images (2,095 subjects) and 4,730 Caucasian male images (2,799 subjects), and 12,383 African-American female images (4,656 subjects) and 12,383 African-American male images (5,414 subjects) for MORPH, and 2,442 Caucasian female images (160 subjects) and 2,442 Caucasian male images (238 subjects) for Notre Dame.


%
%

The difference heatmaps show that the difference in information  between  female and male images is effectively minimized.
Remaining differences at any pixel are generally less than 5\% and are roughly balanced between female and male.
The impostor and genuine distributions for the information-equalized images show a fundamental change from those for the original dataset.
While the impostor distribution for females is still worse than for males,
{\it the genuine distribution for females is now the same as, or slightly better than, the genuine distribution for males.}

Note that this change in the comparison of the female and male genuine distributions occurs solely from creating an approximately information-equalized test set.
The matching algorithm is the same, and was trained on the MS1MV2 dataset which has known female under-representation.
This suggests that the result that females have a worse genuine distribution was, in effect, based on bias in the test data rather than bias in the training data. 

\section{Impostor Difference Reflects Between-Subject Variation}
\label{sec:unequal}
The information-equalized dataset shows that females have a genuine distribution as good or better than males, but
the impostor distribution for females is still worse.
To explore how and why this can be the case, we employ a concept that was once ubiquitous in face recognition, but that may be less known to those entering the field since the wave of deep CNNs - ``eigenfaces'' based on principal component analysis (PCA) \cite{Turk1991, Turk2001}.
The essence of eigenfaces is to learn, from a set of training images,
a face space that accounts for the most variation possible 
in a linear combination of orthogonal dimensions.
While eigenfaces have been used for gender classification (e.g., \cite{Valentin1997}), 
we are not aware of any previous comparison of the dimensionality of face spaces computed separately for females and males.

We computed the face space separately for females and for males,
using one image with the highest fraction of pixels labeled as face for each person, with skin pixels outside of a 10\% mask zeroed.
For simplicity of comparison, 
we randomly selected the same number of males as there are females.
This resulted in 2,798 images each for MORPH Caucasian females and males, 5,929 images each for MORPH African-American females and males, and 169 images each for Notre Dame Caucasian.

\begin{figure*}[t]
  \begin{subfigure}[b]{1\linewidth}
      \begin{subfigure}[b]{0.32\linewidth}
        \centering
          \includegraphics[width=\linewidth]{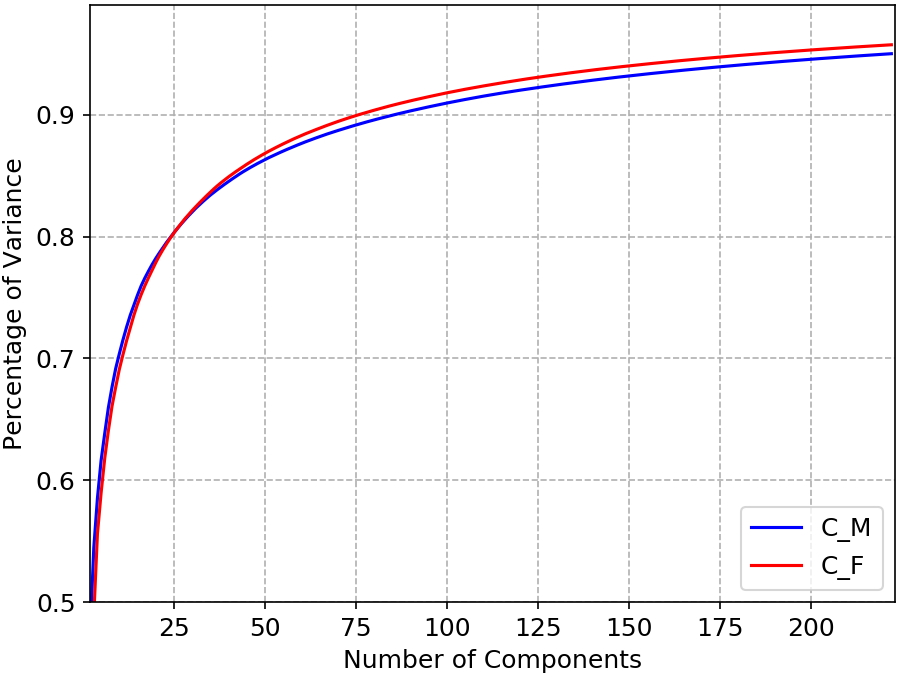}
          \caption{MORPH Caucasian}
      \end{subfigure}
      \hfill 
      \begin{subfigure}[b]{0.32\linewidth}
        \centering
          \includegraphics[width=\linewidth]{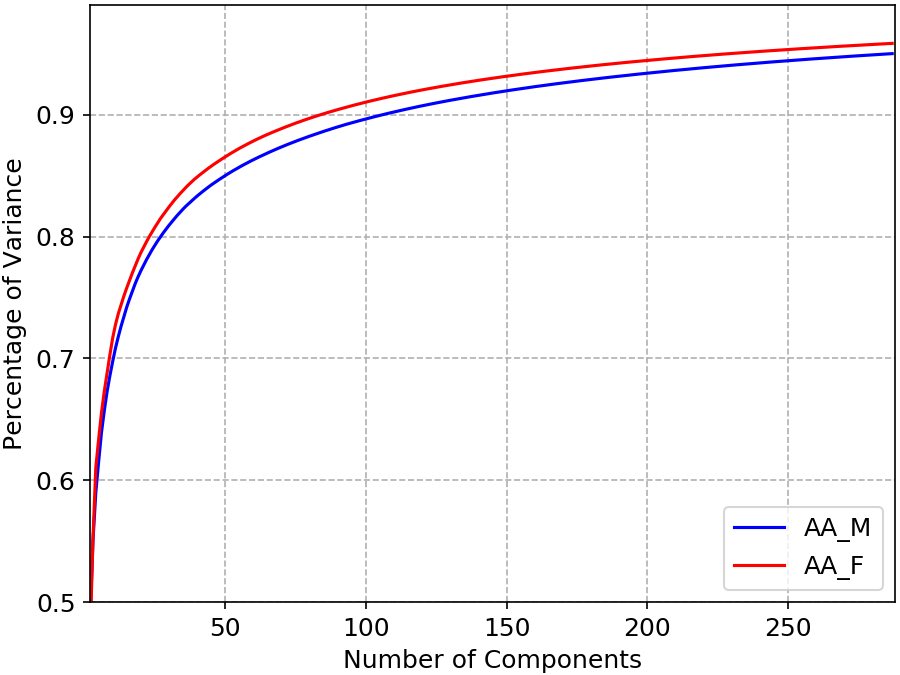}
          \caption{MORPH African-American}
      \end{subfigure}
      \hfill 
      \begin{subfigure}[b]{0.32\linewidth}
        \centering
          \includegraphics[width=\linewidth]{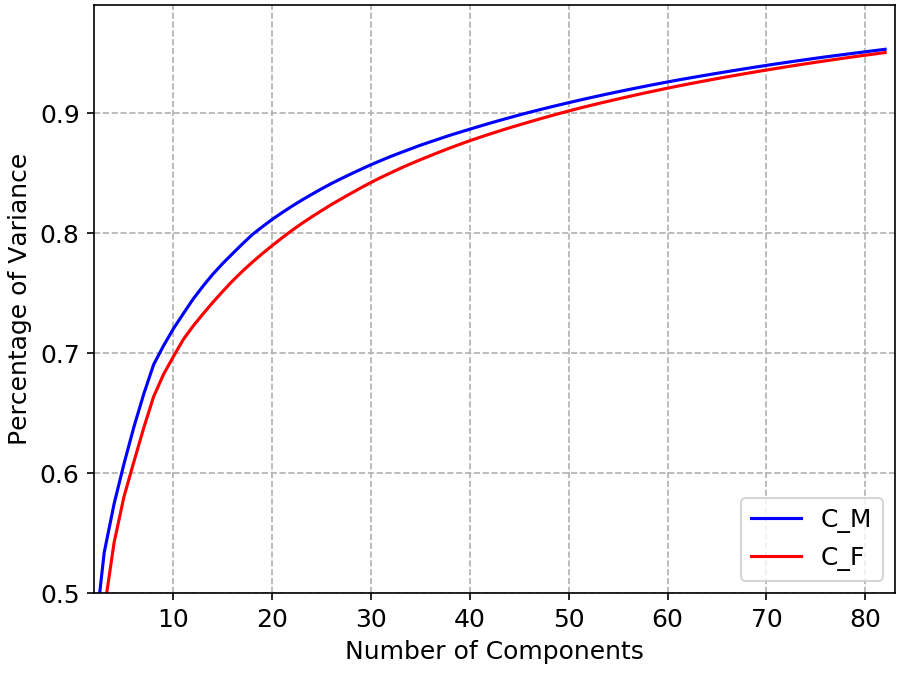}
          \caption{Notre Dame Caucasian}
      \end{subfigure}
  \end{subfigure}
  \caption{PCA on eigenfaces for males and females up to 95\% variance.}
  \label{fig:eigenfaces}
\end{figure*}

Figure \ref{fig:eigenfaces} compares the face space for females and males in terms of the percent of variation accounted for by a given number of eigenfaces.
For MORPH Caucasian,
females need 187 eigenfaces to capture 95\% of variation, whereas males need 222 eigenfaces.
Thus, 95\% of the variation in face images of different persons
is captured in a face space of about 10\% lower dimension for females than for males.
For MORPH African-American, 
females need 229 eigenfaces to capture 95\% of the variation,
whereas African-American males need 287 eigenfaces.
In this case, 95\% of the variation in face images between different persons
is captured in a face space of about 20\% lower dimension for females.
For Notre Dame Caucasian, the result is different, as females require more eigenfaces than males, 82 to 80.
The reason for this difference is not immediately obvious,
but the Notre Dame dataset is different in several respects.
One, there are only 169 subjects / images contributing to each face space.
Two, the subjects are overwhelming young, 
with over 90\% of both female and male subjects in the age range 18-29, whereas the large majority of the subjects in the MORPH African-American and MORPH Caucasian are 30 or older.
Three, and perhaps related to the small number of subjects almost all of whom are young, a much smaller number of eigenfaces is needed to reach 95\% of variation. 

The curves in Figure \ref{fig:eigenfaces} are for the central masked face regions.
The results are similar if the original 112x112 images are used.
For example, for MORPH Caucasian, females need 303 eigenfaces and males need 331 to capture 95\% of variance.
Overall, the implication of this face space analysis is that,
for a given dimension of face space, 
images of two different females are more similar than images of two different males.

\begin{figure}[t]
    \centering
      \begin{subfigure}[b]{0.32\linewidth}
        \centering
          \includegraphics[width=\linewidth]{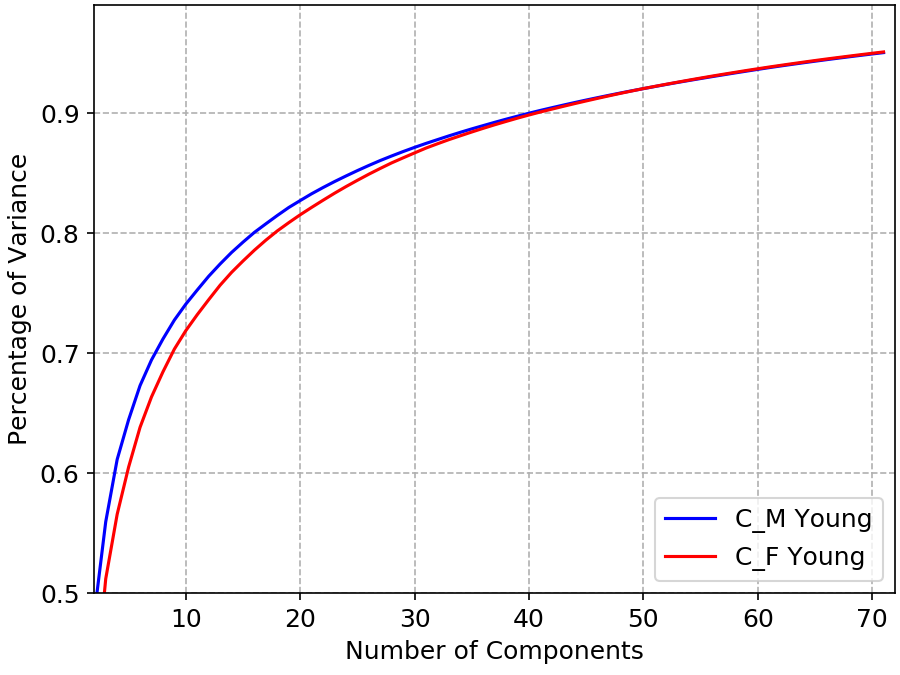}
      \end{subfigure}
  \caption{MORPH Caucasian with same number of subjects and age distribution as Notre Dame dataset.}
  \label{fig:morph_c_match_nd_pca}
\end{figure}

To determine if the size of the Notre Dame dataset is the problem, we create male and female subsets of MORPH Caucasian images, with same number of subjects and same age distribution as the Notre Dame dataset.
Figure \ref{fig:morph_c_match_nd_pca} shows that when this subset of MORPH data is used, the result is similar to the Notre Dame dataset.

\section{Face Spaces and Impostor Distributions}

The previous section shows that, for the same number of subjects and images, the male image set requires a larger dimension of face space than the female image set to account for the same percent variation.
For the same dimensionality of face space, a greater percent of the variation in female face appearance is accounted for.
In face space, faces of two different females are more similar-appearing on average than faces of two different males.
We conjecture that this is an explanation for why the male impostor distribution is centered at a lower similarity value than the female impostor distribution.

\begin{figure*}[t]
  \begin{subfigure}[b]{1\linewidth}
      \begin{subfigure}[b]{1\linewidth}
        \centering
          \begin{subfigure}[b]{0.32\columnwidth}
            \centering
            \includegraphics[width=\linewidth]{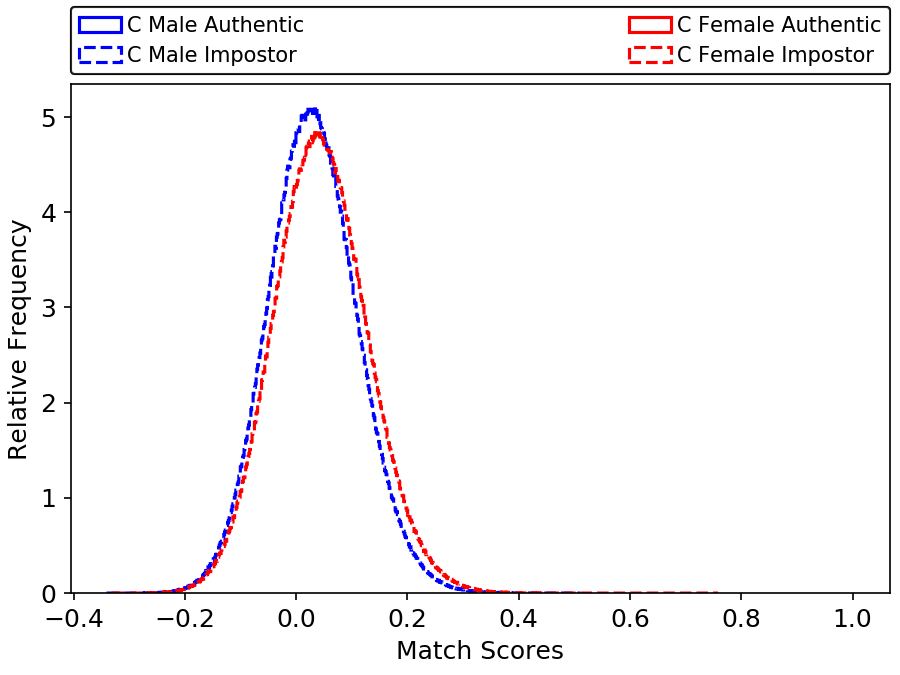}
          \end{subfigure}
          \begin{subfigure}[b]{0.32\columnwidth}
            \centering
            \includegraphics[width=\linewidth]{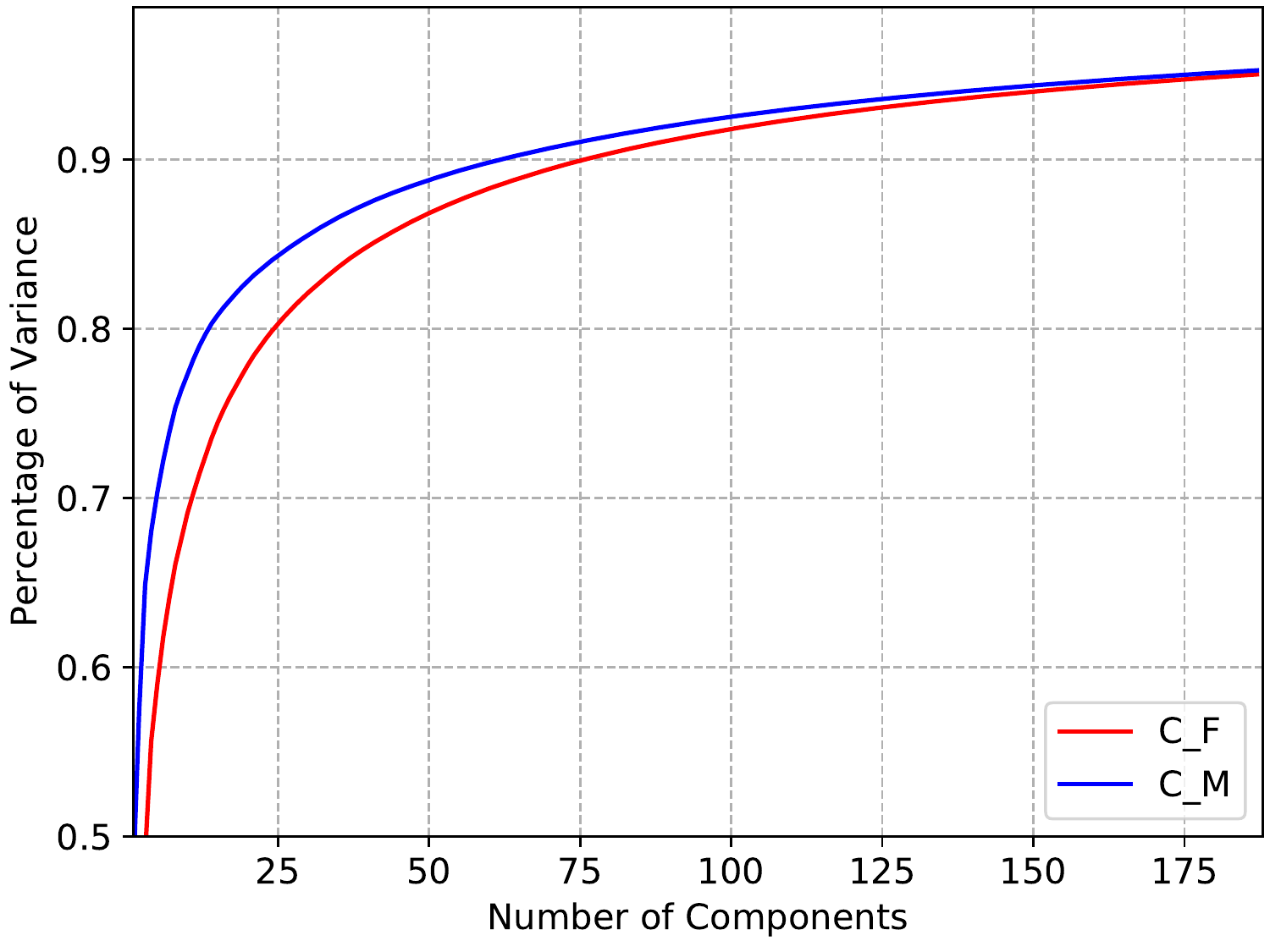}
          \end{subfigure}
          \begin{subfigure}[b]{0.32\columnwidth}
            \centering
            \includegraphics[width=\linewidth]{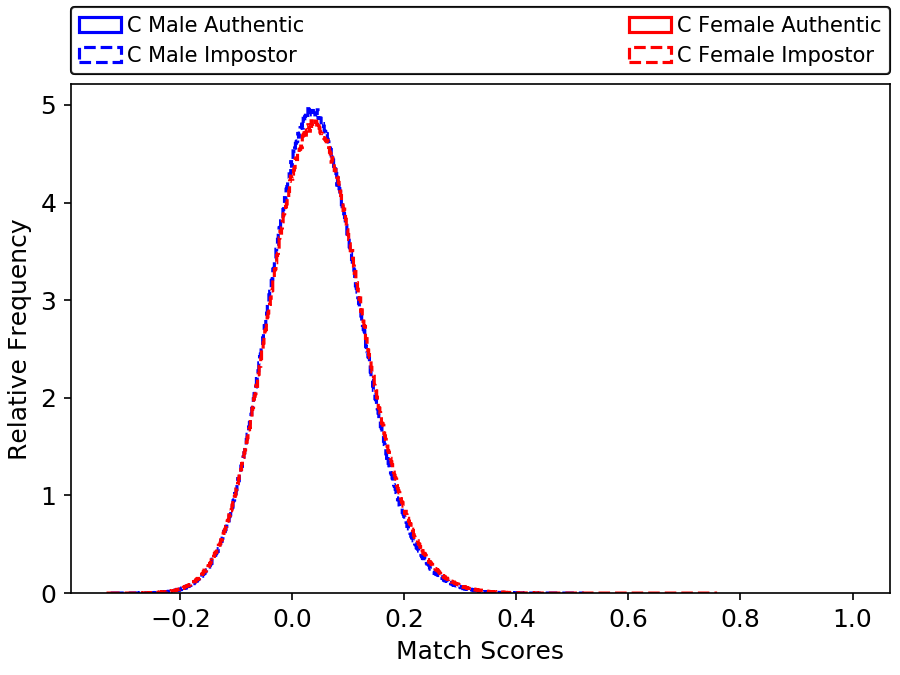}
          \end{subfigure}
          \caption{MORPH Caucasian}
      \end{subfigure}
      \hfill 
      \begin{subfigure}[b]{1\linewidth}
        \centering
          \begin{subfigure}[b]{0.32\columnwidth}
            \centering
            \includegraphics[width=\linewidth]{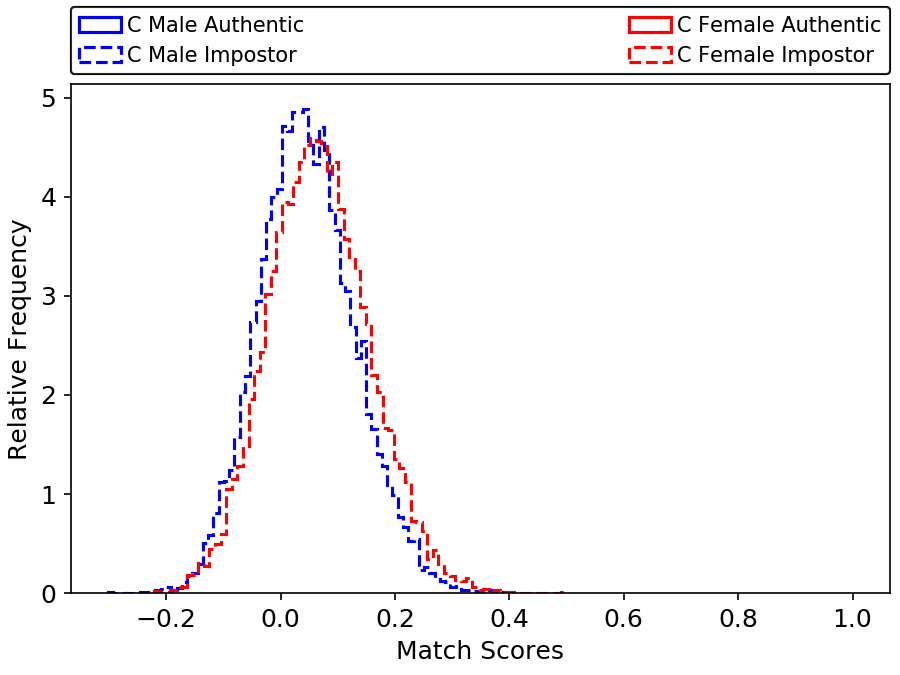}
          \end{subfigure}
          \begin{subfigure}[b]{0.32\columnwidth}
            \centering
            \includegraphics[width=\linewidth]{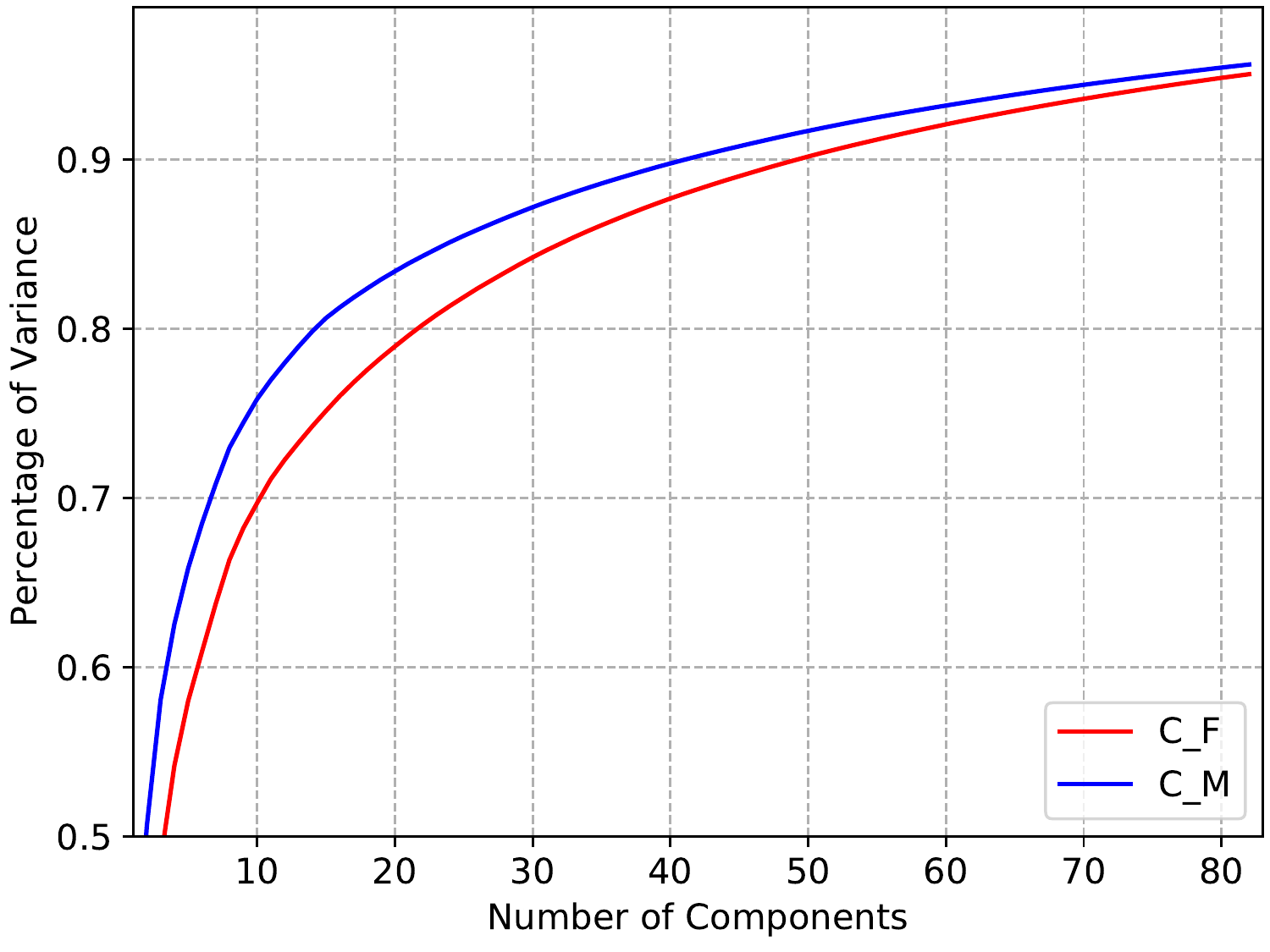}
          \end{subfigure}
          \begin{subfigure}[b]{0.32\columnwidth}
            \centering
            \includegraphics[width=\linewidth]{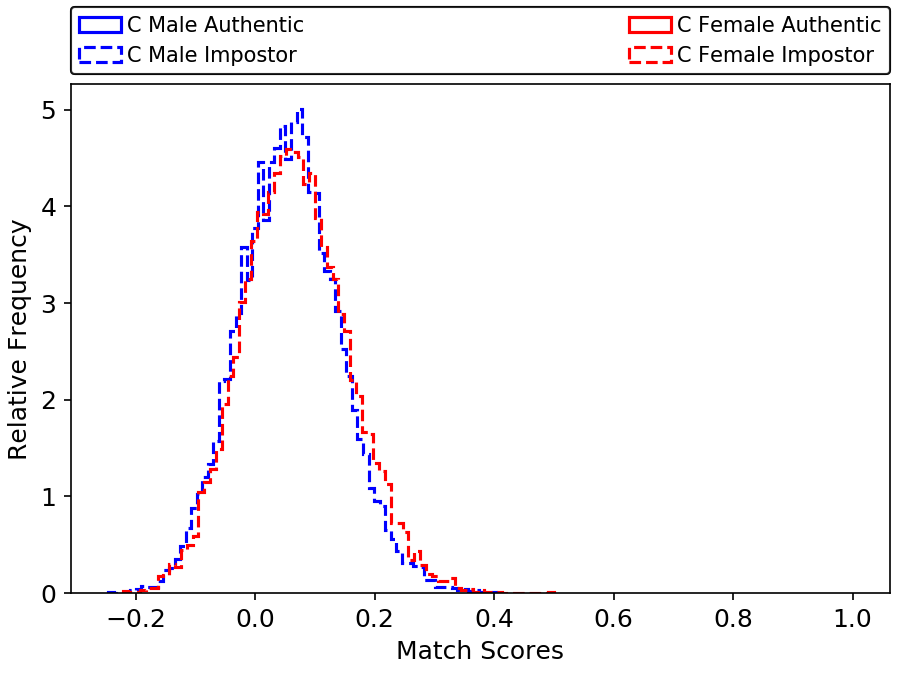}
          \end{subfigure}
          \caption{Notre Dame Caucasian}
      \end{subfigure}
  \end{subfigure}
  \caption{PCA analysis after removing male images with error at 80\% variance (middle), and the impostor distribution with random matching (left) and with 80\% removal (right).}
  \label{fig:pca_removal}
\end{figure*}

To investigate this further, we select a subset of male face images that generates a face space that better matches that of the female image set.
We select the same number of males as there are females, and use one image per person, and we select the male images that have the lowest reconstruction error at 80\% variance. 
This selection flipped the relative position of the female and male PCA curves as seen in Figure \ref{fig:pca_removal}.
We then compare the impostor distribution of the males random selection and the 80\% variance selection.
Figure \ref{fig:pca_removal} shows that the impostor distribution for males and females is closer when the PCA curve is flipped, meaning that when males have less variation in between-subject, their impostor matches are similar to women.

We were able to reproduce the PCA inverting experiment with the MORPH Caucasian and Notre Dame dataset, but not on MORPH African-American.
The original difference in the PCA between African-American males and females is larger than the other two datasets, and when we remove male images with highest error rate, females still require less components to achieve the same variation.
This issue exists even if we reduce womens' data to half, so that we can have more male data to remove, achieving a much closer PCA curve, but that does not flip and does not come as close as the MORPH Caucasian and Notre Dame dataset do.
In our experiments, to achieve a more similar impostor distribution across gender, the males' PCA curve must be higher (require less components) than the females, if they are near-equal, male impostor distribution is still better.

\section{Conclusions and Discussion}

The results in Figure \ref{fig:auth_imp} are representative of the consensus of previous research that face recognition accuracy is lower for females. 
The most extensive study in this area, in terms of number of different matchers and datasets considered, is the recent NIST report on demographic effects \cite{frvt3}.
This report found that females consistently have higher FMR (worse impostor distribution), and generally also have higher FNMR (worse genuine distribution) but that there are exceptions to the higher FNMR generalization.

However, the studies covered in Related Work are done with test sets that are largely uncontrolled with respect to differences between female and male images.
One recommendation for future studies on demographic differences in face recognition accuracy is to include information such as the \% face distributions or the face difference heatmap for the test images.
This would encourage an awareness of how test data may come with a built-in ``bias'' toward observing certain accuracy differences.

Female / male differences in size and shape of the face are well known and have long been studied in diverse fields 
such as ergonomics \cite{Zhuang2010}, anthropology \cite{Holton2014}, and surgery \cite{Farkas2005}.
%
%
Female and male faces also change differently with age \cite{Albert2007}.
As a matter of biology, the range of size and shape of the female face is smaller than the range of size and shape of the male face.
In the face recognition context, 
this becomes evident in an eigenface comparison of the female and male ``face spaces''.

Thus, an initial condition for face matchers is the fact that, on average, two different female faces actually do appear more similar than two different male faces.
The impostor distribution for females is naturally centered at higher similarity values that the impostor distribution for males.
Any face matching algorithm that computes a similarity score without, in effect, classifying the face as female or male and allowing that to influence the score, seems likely to have different female / male impostor distributions.

The range of face size and shape being smaller for females can also cause the distribution of genuine scores for females to be centered at higher similarity values than for males.
This is the result seen for the information-equalized dataset in Figure \ref{fig:equalized_images}.
The fact that the opposite is commonly found in female / male comparisons of accuracy, with the female genuine distribution worse than the male, is due to effects of gendered hairstyles combined with biological differences in face size and shape.
Most of the difference in the \% face distributions in Figure \ref{fig:skin_dist} is due to the effects of gendered hairstyles.

Is face recognition technology sexist?
The worse genuine distribution seems due primarily to the gender differences in hairstyles.
The worse impostor distribution seems due primarily to the biological differences in the range of female and male face appearance.
So, the ``sexist'' result of lower accuracy for females seems largely due to a combination of biological differences and gendered hairstyle conventions, and not to face recognition methods and training datasets.

Future work could include analyzing the female/male difference for an Asian image cohort, or other cohorts.

\section*{Acknowledgment}
The authors would like to thank Ross Beveridge, Alice O'Toole and Volker Blanz for helpful discussions on this project.

\bibliography{main.bib}

\end{document}